\documentclass{article}

\PassOptionsToPackage{numbers, sort}{natbib}

\usepackage{graphicx}
\usepackage{subcaption}
\usepackage{amsmath}
\usepackage{lipsum}

\usepackage{afterpage}
\usepackage{placeins}  %

\usepackage{multirow}
\usepackage[export]{adjustbox}  %

 \usepackage{import}  %

\usepackage{xspace}
\def\eg{\emph{e.g.}\@\xspace}
\def\ie{\emph{i.e.}\@\xspace}
\def\etal{\emph{et al.}\@\xspace}
\def\cf{\emph{c.f.}\@\xspace}
\def\etc{\emph{etc}}
\newcommand{\bd}[1]{{{\bf #1}}}
\newcommand{\st}[1]{{{\scriptsize #1}}}
\newcommand{\rotintab}[1]{\rotatebox[origin=c]{90}{#1}}

\renewcommand{\paragraph}[1]{\smallskip\noindent{\bf{#1}}}

\DeclareMathOperator*{\argmax}{arg\,max}
\newcommand{\loss}{\mathcal{L}}

\def\NeRFre{NeRF$^{\dagger}$\@\xspace}

\def\NeRFour{NeRF-ID\@\xspace}

\newcommand{\capNerf}{
NeRF denotes the results reported in the original paper~\cite{Mildenhall20},
\NeRFre is our reimplementation.
}

\newcommand{\isArXiv}[2]{#1}
\isArXiv{
\usepackage[preprint]{neurips_2021}
}{
\usepackage{neurips_2021}
}

\usepackage[utf8]{inputenc} %
\usepackage[T1]{fontenc}    %
\usepackage{hyperref}       %
\usepackage{url}            %
\usepackage{booktabs}       %
\usepackage{amsfonts}       %
\usepackage{nicefrac}       %
\usepackage{microtype}      %
\usepackage{xcolor}         %

\isArXiv{
\hypersetup{pdfauthor={Relja Arandjelovic},pdftitle={NeRF in detail: Learning to sample for view synthesis}}
}{
}

\title{NeRF in detail: Learning to sample for view synthesis}

\author{%
	\textbf{Relja Arandjelovi\'c}\textsuperscript{1}
	\quad
	\textbf{Andrew Zisserman}\textsuperscript{1,2}
	\\\\
	$^1$DeepMind \quad {$^2$VGG, Dept.\  of Engineering Science, University of Oxford}
}

\newcommand{\figNerfGen}{
\def\OverviewH{6cm}
\def\ProposalH{4cm}
\begin{figure}[t]
\centering
\hspace*{-1cm}
\begin{tabular}{ccc}
\multirow{3}{*}{
\includegraphics[height=\OverviewH,valign=c]{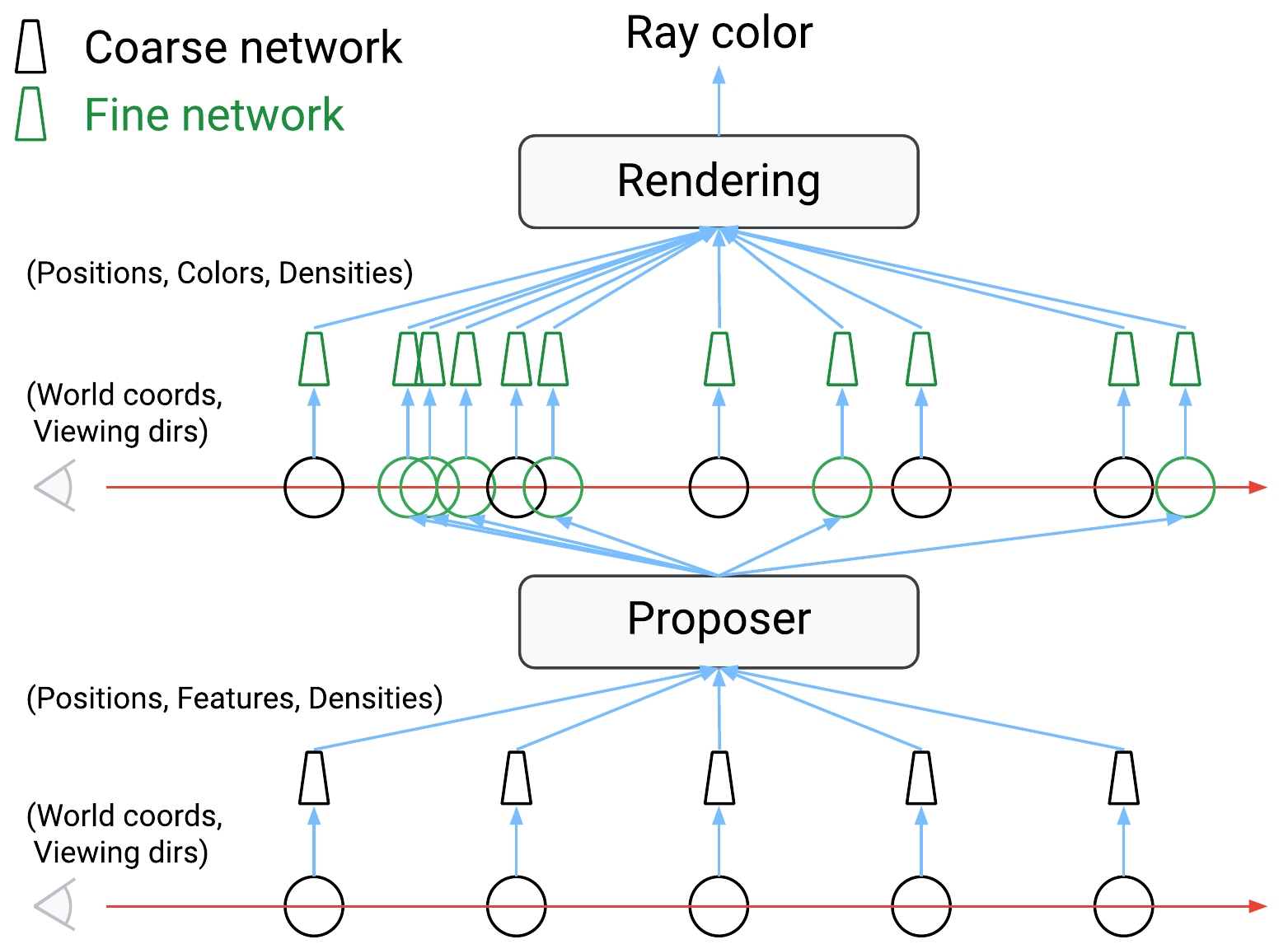}
} &
\includegraphics[height=\ProposalH,valign=c]{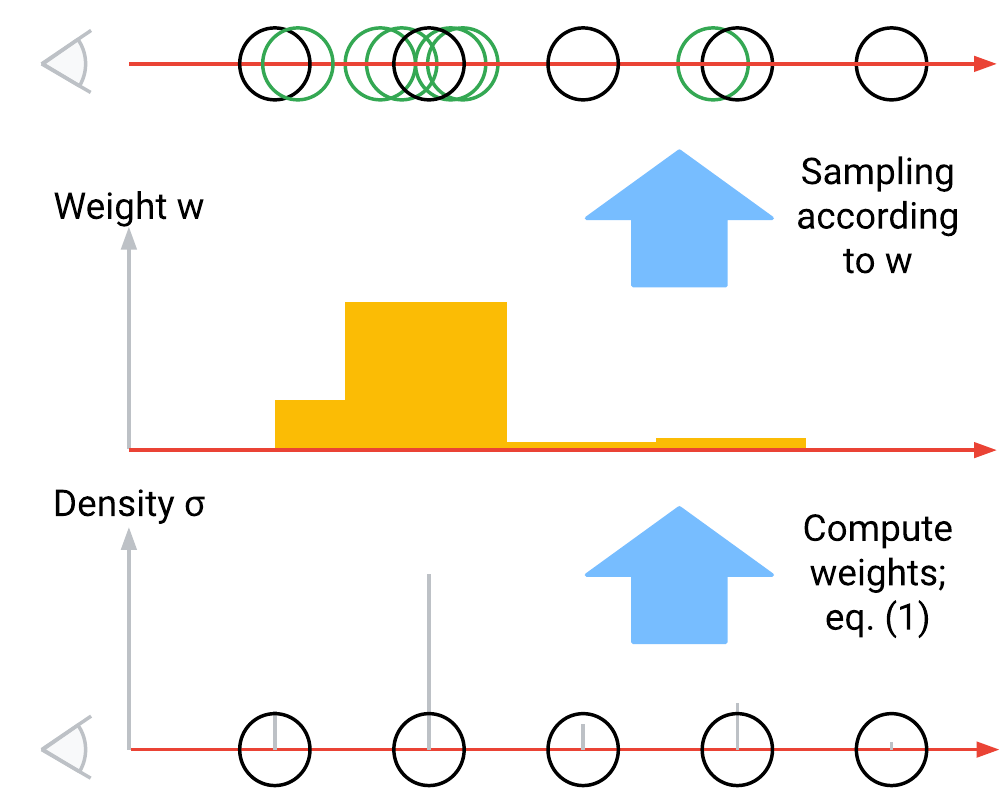} \\
&
(b) Heuristic proposer (vanilla NeRF) \\
\addlinespace[1em]
&
\includegraphics[height=\ProposalH,valign=c]{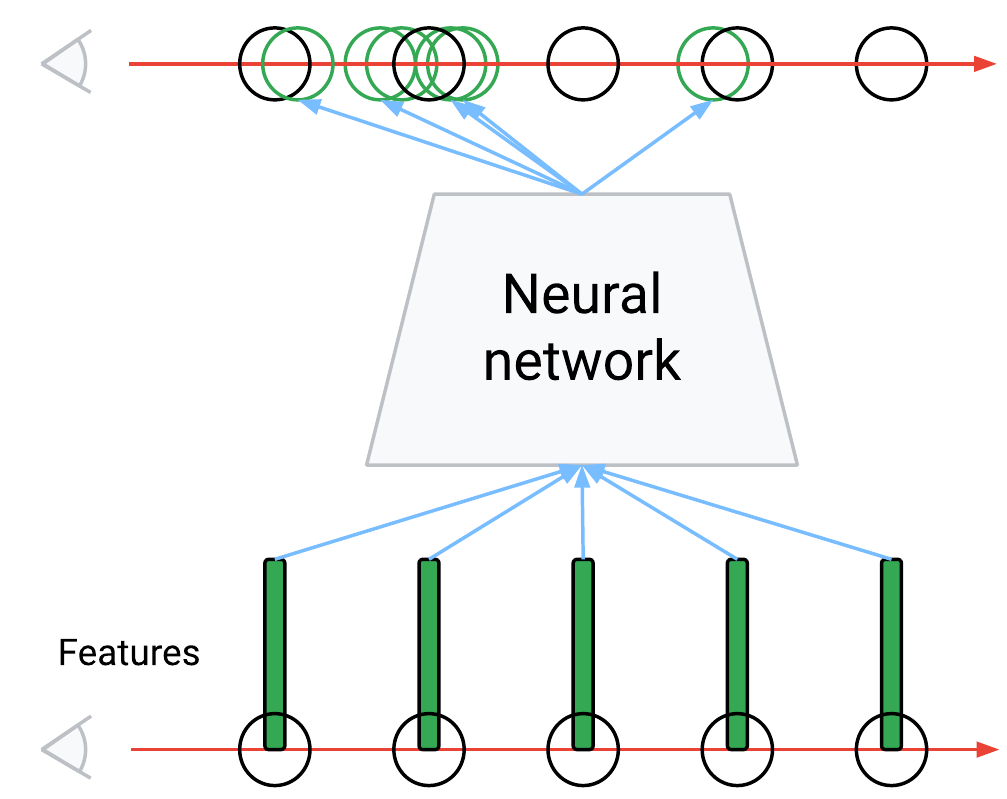} \\
(a) NeRF's coarse-to-fine mechanism &
(c) Learnt proposer (\NeRFour, this work)
\end{tabular}
\caption{{\bf Overview of NeRF and our method.}
NeRF's coarse-to-fine approach (a)
relies on a heuristic `proposer' (b) which
acts on the output of the coarse network
and produces samples to pass to the fine network.
We substitute this mechanism with a learnable proposer (c).
}
\label{fig:nerfgen}
\end{figure}
}

\newcommand{\figProposer}{
\def\ProposerH{4cm}
\begin{figure}[t]
\hspace*{-1.7cm}
\centering
\begin{tabular}{ccc}
\includegraphics[height=\ProposerH,valign=c]{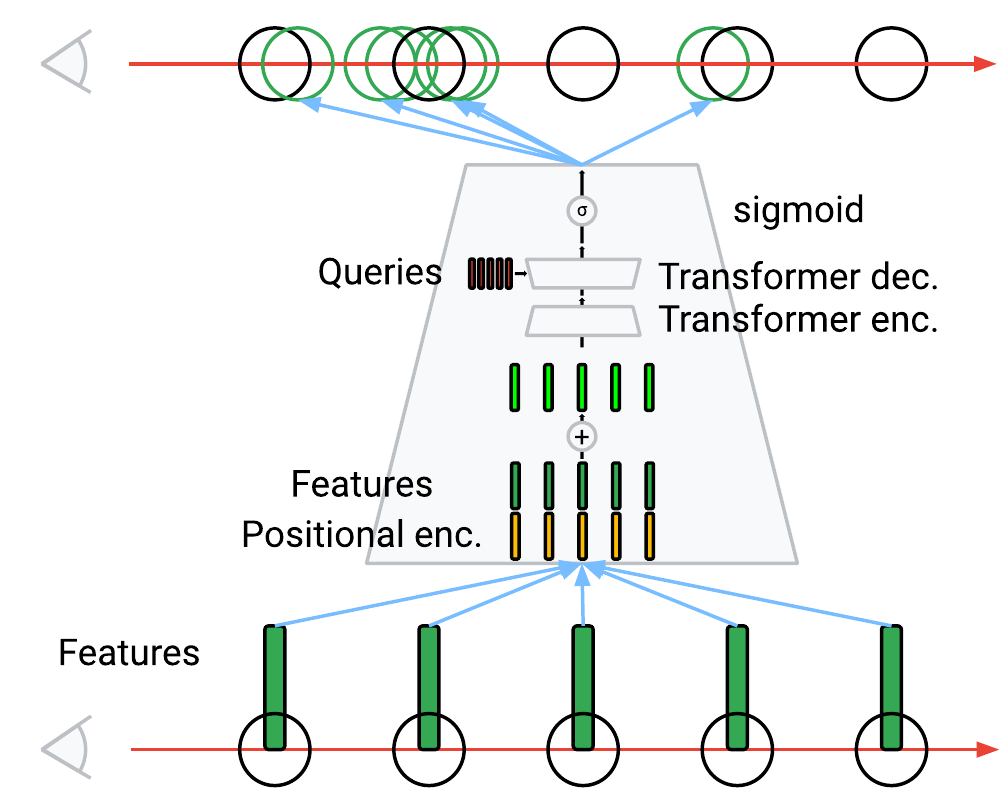}
&
\includegraphics[height=\ProposerH,valign=c]{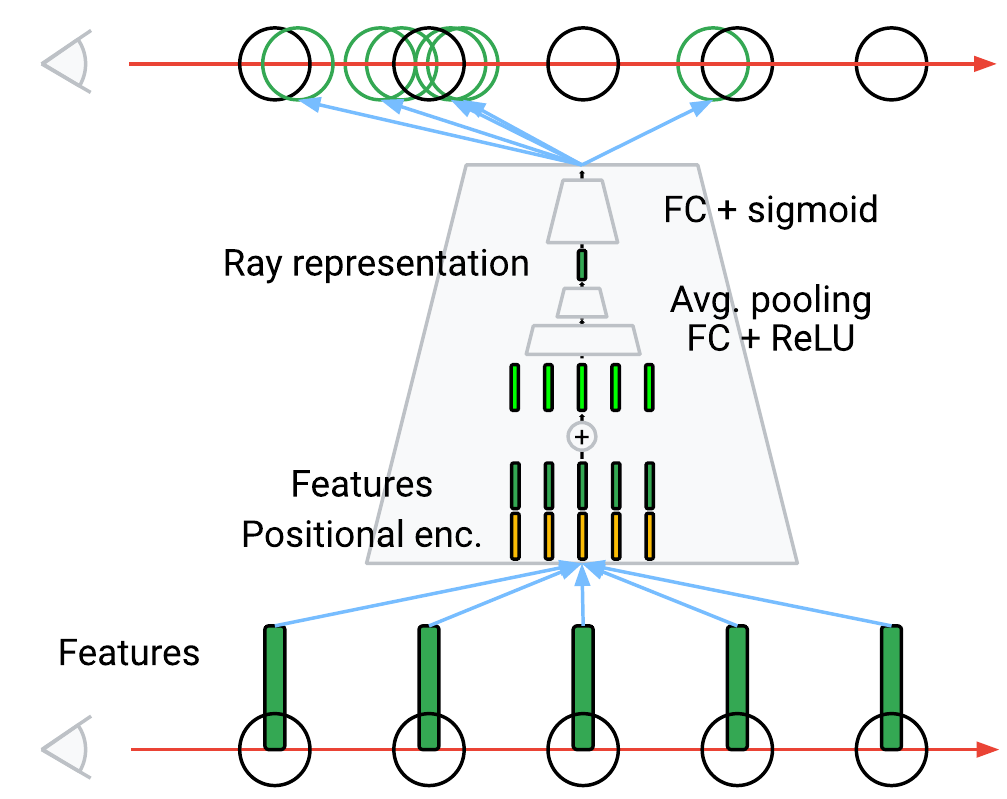}
&
\includegraphics[height=\ProposerH,valign=c]{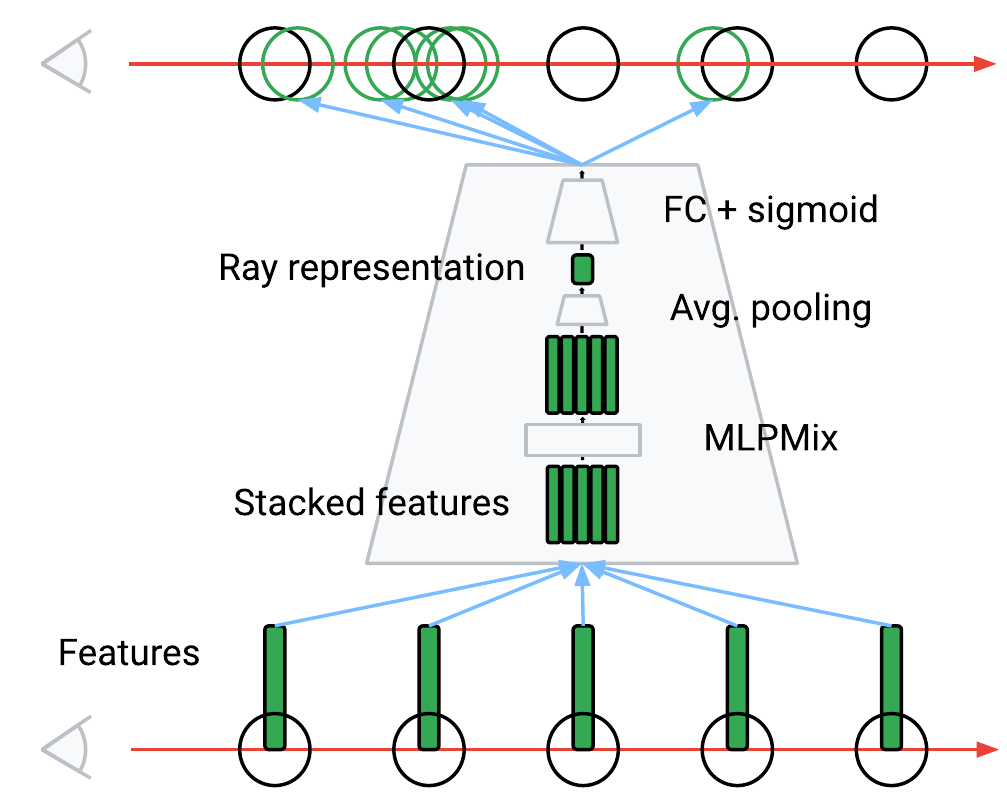}
\\
(a) Transformer &
(b) Pool &
(c) MLPMix
\\
\end{tabular}
\caption{{\bf Trainable proposer architectures.}
The input features come from the course network,
as shown in Figures~\ref{fig:nerfgen}(a) and (c).
Full details are in \isArXiv{Appendix~\ref{sec:app:arch} and Figure~\ref{fig:archdetail}}{the supplementary material}.
}
\label{fig:proposer}
\end{figure}
}

\newcommand{\figResQual}{
\def\QualH{2.7cm}
\begin{figure}[t]
\centering
\begin{tabular}{ccccc}
\rotintab{Blender-Lego} &
  \includegraphics[height=\QualH,valign=c]{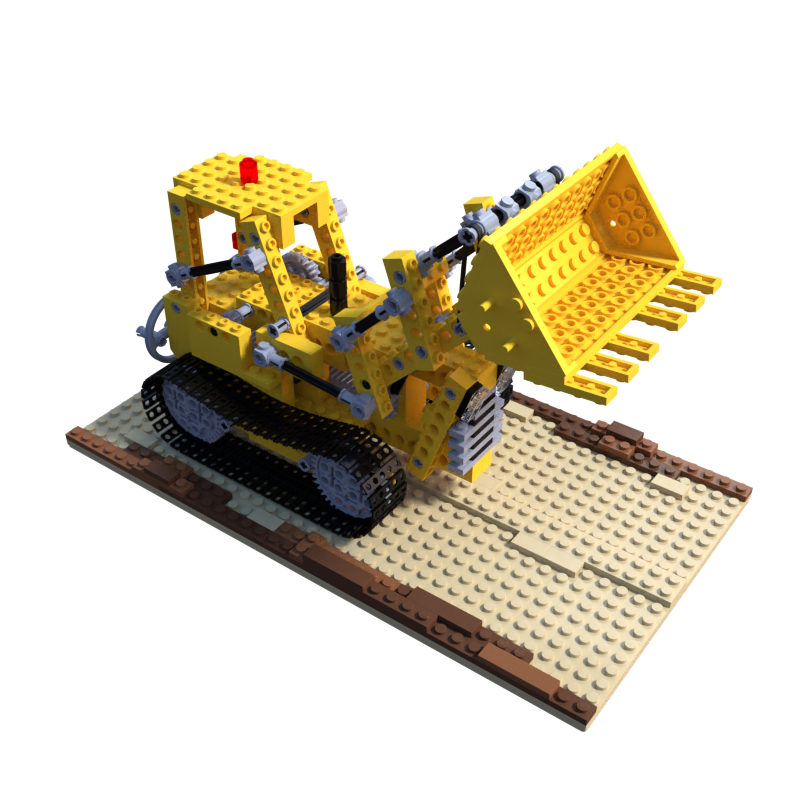} &
  \includegraphics[height=\QualH,valign=c]{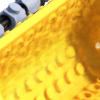} &
  \includegraphics[height=\QualH,valign=c]{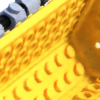} &
  \includegraphics[height=\QualH,valign=c]{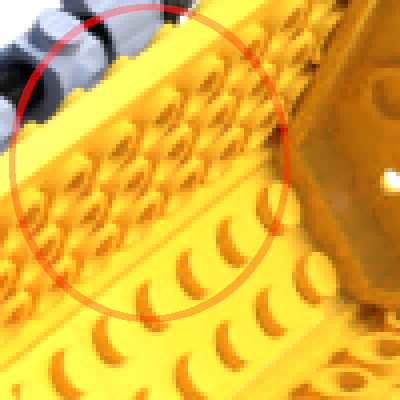}
\\ \addlinespace[0.5em]
\rotintab{Blender-Ficus} &
  \includegraphics[height=\QualH,valign=c]{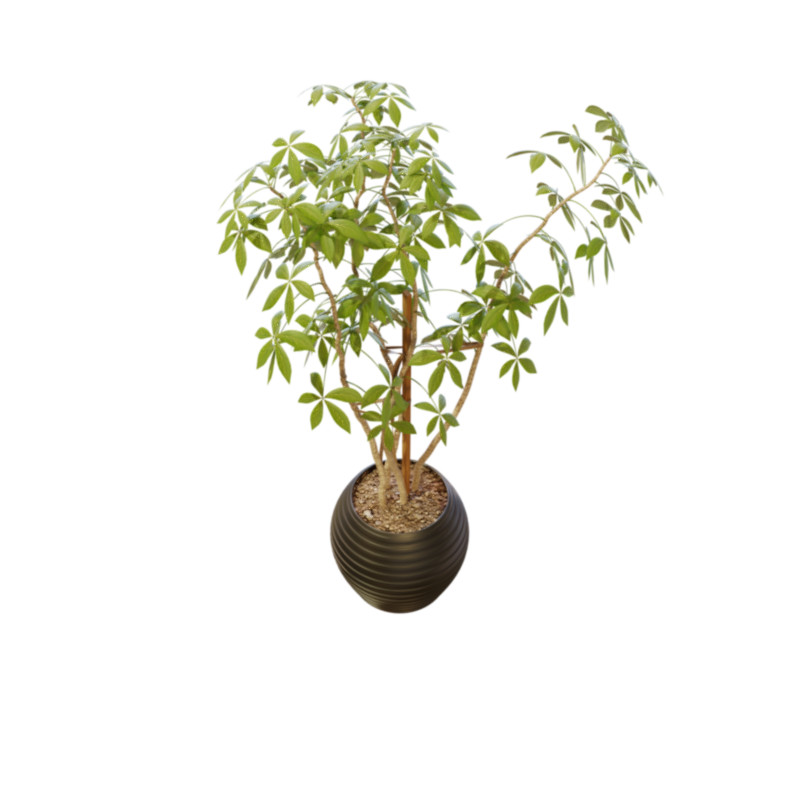} &
  \includegraphics[height=\QualH,valign=c]{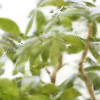} &
  \includegraphics[height=\QualH,valign=c]{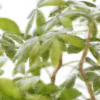} &
  \includegraphics[height=\QualH,valign=c]{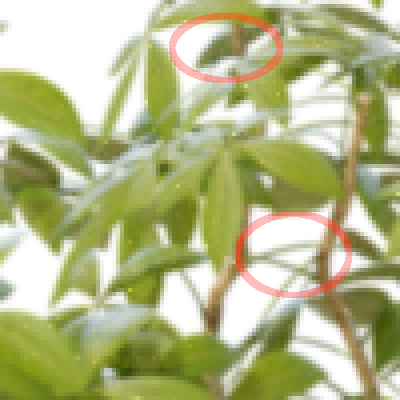}
\\ \addlinespace[0.5em]
\rotintab{Blender-Chair} &
  \includegraphics[height=\QualH,valign=c]{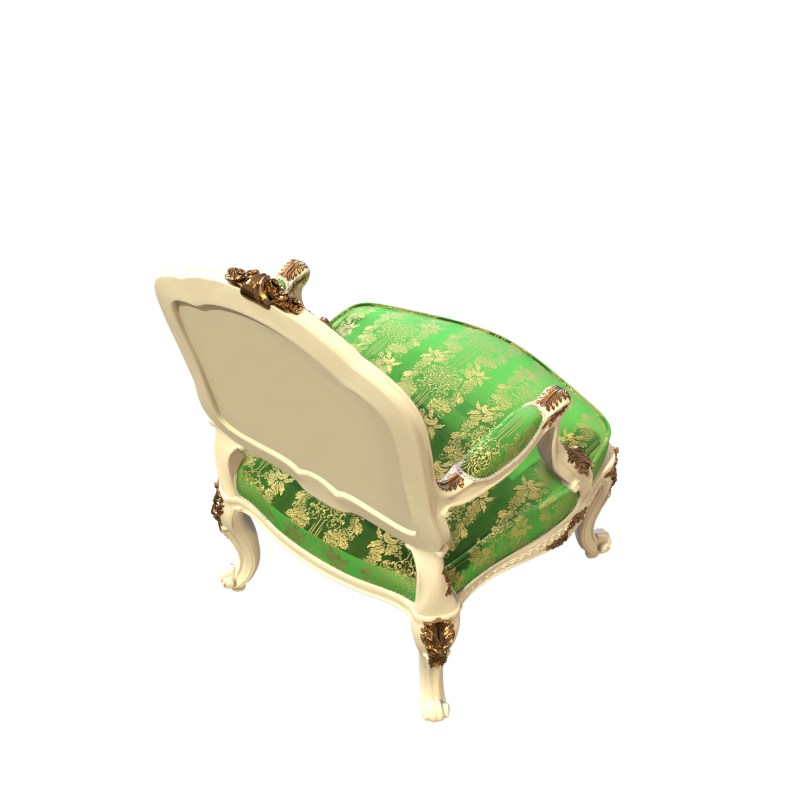} &
  \includegraphics[height=\QualH,valign=c]{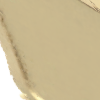} &
  \includegraphics[height=\QualH,valign=c]{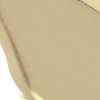} &
  \includegraphics[height=\QualH,valign=c]{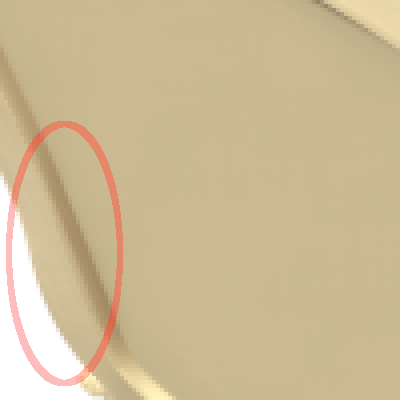}
\\ \addlinespace[0.5em]
\rotintab{Blender-Ship} &
  \includegraphics[height=\QualH,valign=c]{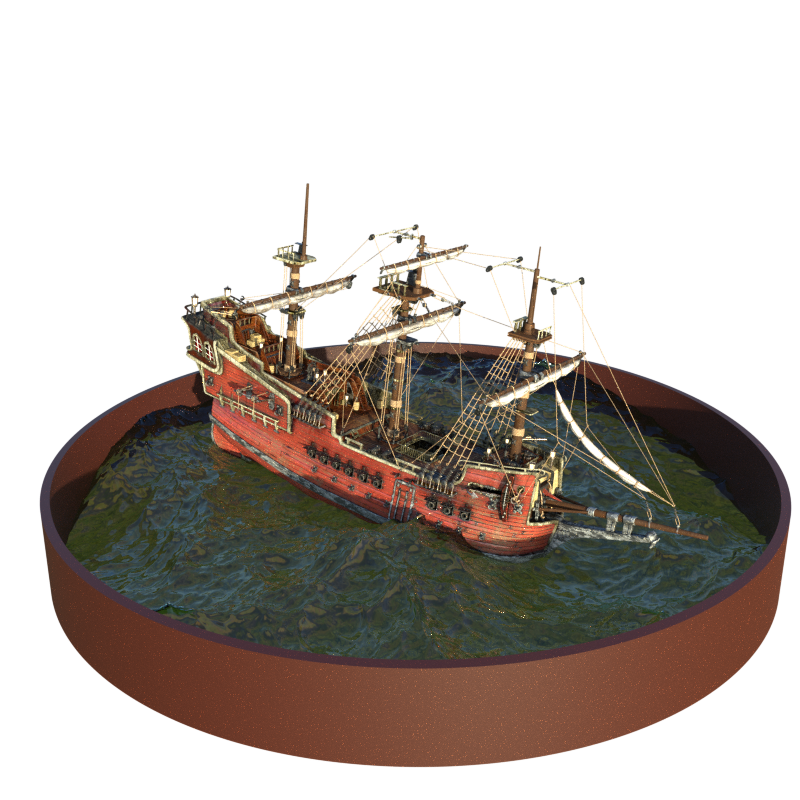} &
  \includegraphics[height=\QualH,valign=c]{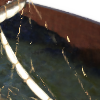} &
  \includegraphics[height=\QualH,valign=c]{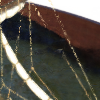} &
  \includegraphics[height=\QualH,valign=c]{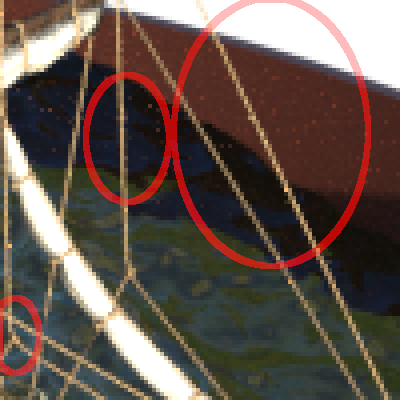}
\\ \addlinespace[0.5em]
\rotintab{LLFF-TRex} &
  \includegraphics[height=\QualH,valign=c]{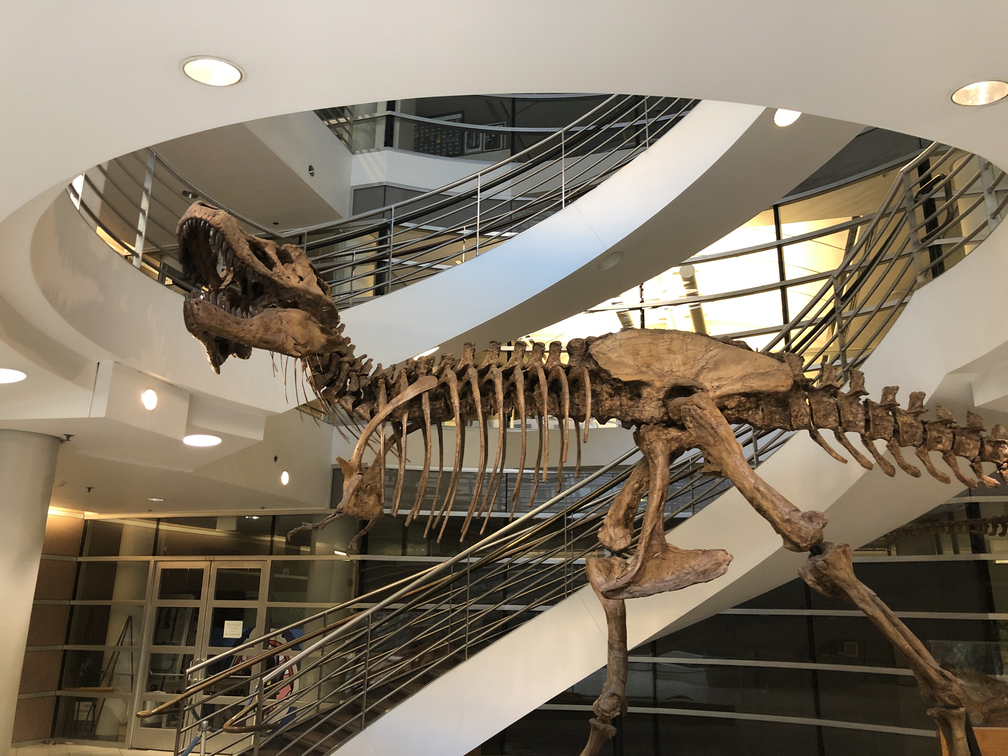} &
  \includegraphics[height=\QualH,valign=c]{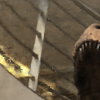} &
  \includegraphics[height=\QualH,valign=c]{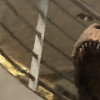} &
  \includegraphics[height=\QualH,valign=c]{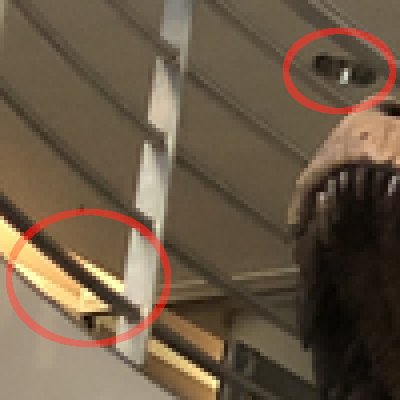}
\\ \addlinespace[0.5em]
\rotintab{LLFF-Fern} &
  \includegraphics[height=\QualH,valign=c]{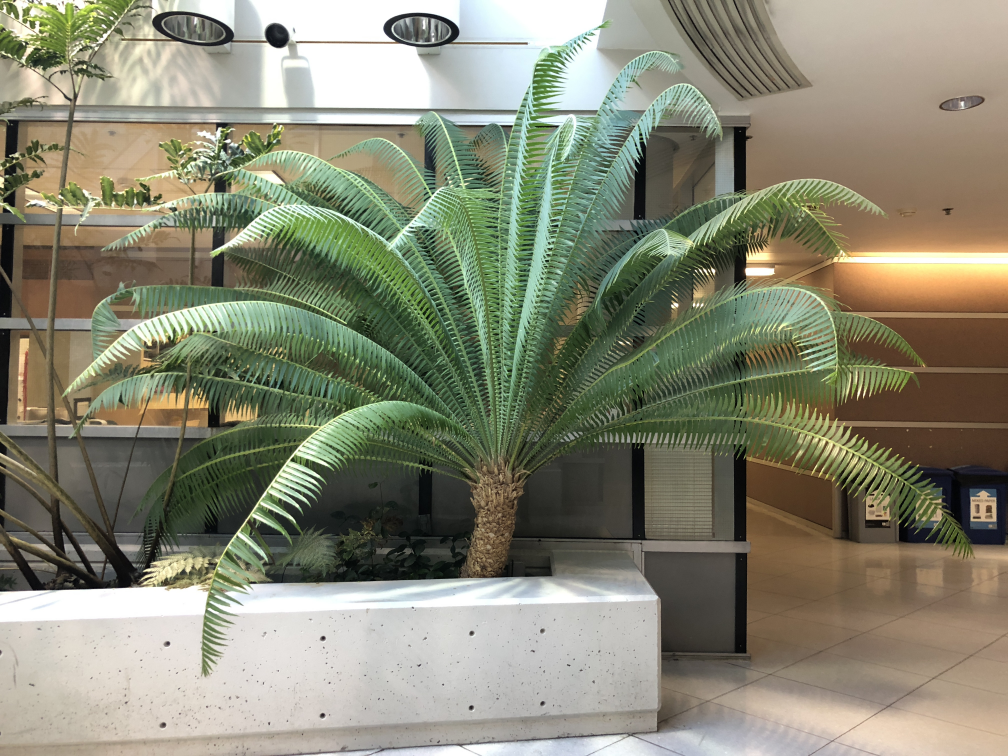} &
  \includegraphics[height=\QualH,valign=c]{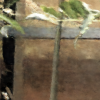} &
  \includegraphics[height=\QualH,valign=c]{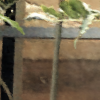} &
  \includegraphics[height=\QualH,valign=c]{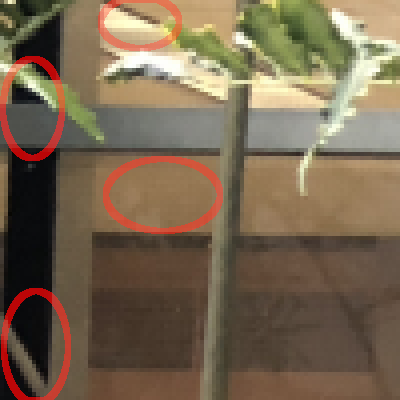}
\\ \addlinespace[0.5em]
\rotintab{LLFF-Horns} &
  \includegraphics[height=\QualH,valign=c]{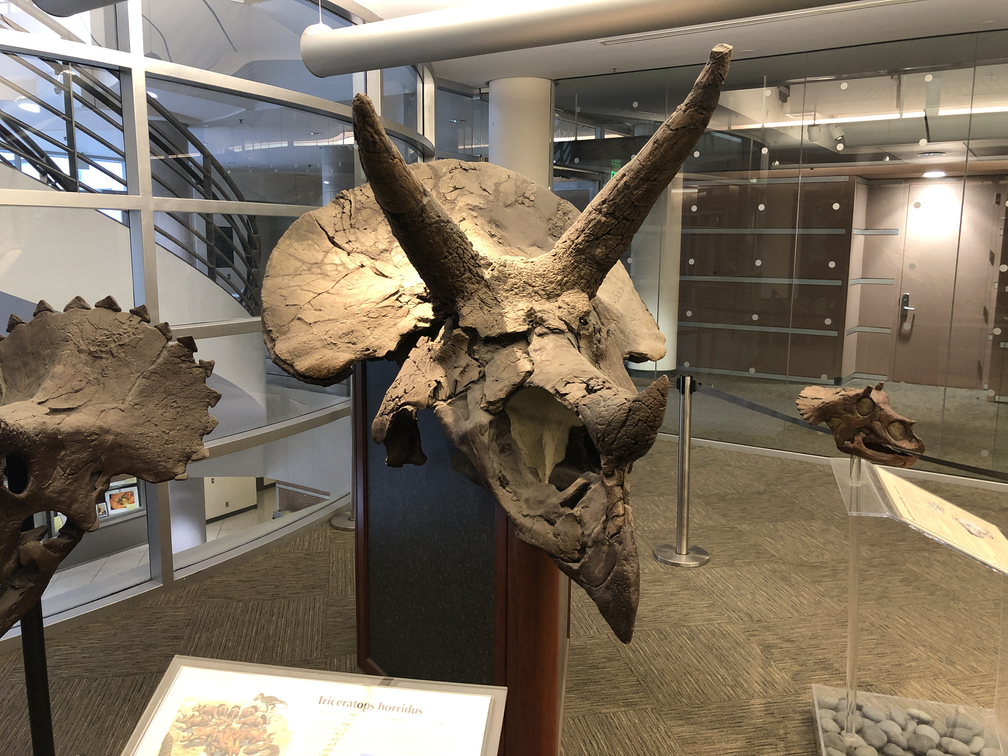} &
  \includegraphics[height=\QualH,valign=c]{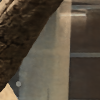} &
  \includegraphics[height=\QualH,valign=c]{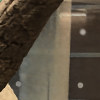} &
  \includegraphics[height=\QualH,valign=c]{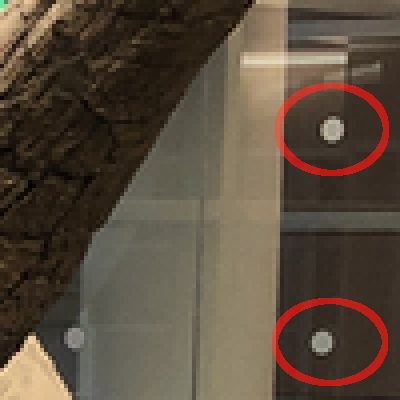}
\\ \addlinespace[0.5em]
& Test set image & \NeRFre & NeRF-ID (Ours) & {\shortstack{Ground truth \\ with highlights}}
\end{tabular}
\caption{{\bf Qualitative results.}
Comparison of our NeRF-ID versus \NeRFre.
Overall both methods produce good renderings, but the difference is especially
apparent in fine details that \NeRFre often misses while NeRF-ID reproduces better,
such as thin branches, ropes, markings, edges \etc.
}
\label{fig:res:qual}
\end{figure}
}

\newcommand{\figSpeedupLlff}{
\begin{figure}[t]
\centering
\includegraphics[width=0.9\linewidth]{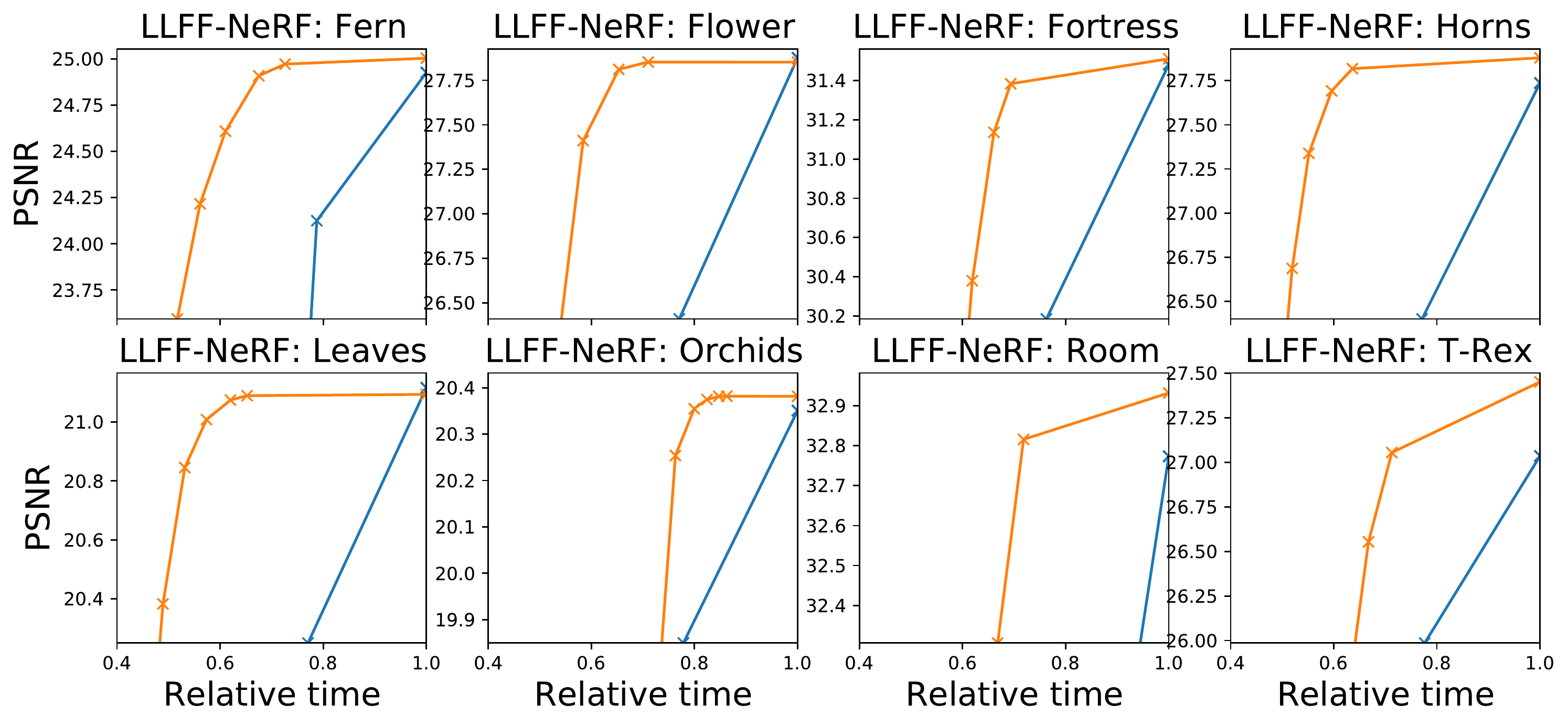}
\caption{{\bf Speedup via importance prediction,
{\color{blue} \NeRFre} vs.\ our {\color{orange} \NeRFour}.}
Samples deemed to be important by the proposer are kept,
different operating points are obtained by varying the importance threshold.
`Relative time' is the time spent on rendering the scene
relative to using all samples.
}
\label{fig:res:speedupllff}
\end{figure}
}

\newcommand{\tabArch}{
\begin{table}[t]
\centering
\caption{{\bf Comparison of proposer architectures and vanilla NeRF (PSNR).}
Most experiments were repeated 5 times and we report the median PSNR,
full results are in \isArXiv{Appendix~\ref{sec:app:results}}{the supplementary}.
\capNerf
Highlighted are the \bd{best} and \underline{second best} results.}
\hspace*{-1cm}
\begin{tabular}{clccccccccc} \toprule
\multirow{6}{*}{\rotintab{\emph{Blender}}} &
Method & Avg. & Chair & Drums & Ficus & Hotdog & Lego & Materials & Mic & Ship \\
\cmidrule(lr){2-2} \cmidrule(lr){3-3} \cmidrule(lr){4-11}
&
NeRF~\cite{Mildenhall20} &  31.01 &  33.00 & 25.01 & 30.13 & 36.18 & 32.54 & 29.62 & 32.91 & 28.65  \\
&
\NeRFre &  31.83 &  34.35 & 24.98 & 30.40 & 37.04 & 33.56 & 30.19 & 34.60 & 29.52  \\
&
Transformer &  31.82 &  \underline{34.50} & 24.91 & 30.93 & 36.78 & 33.84 & \underline{30.27} & \underline{34.57} & 28.75  \\
&
Pool &  \underline{32.23} &  34.28 & \bd{25.18} & \bd{32.37} & \underline{37.11} & \bd{34.74} & 29.92 & 34.31 & \bd{29.94}  \\
&
MLPMix &  \bd{32.34} &  \bd{34.54} & \underline{25.15} & \underline{32.24} & \bd{37.26} & \underline{34.73} & \bd{30.37} & \bd{34.71} & \underline{29.75}  \\

\midrule

\multirow{6}{*}{\rotintab{\emph{LLFF-NeRF}}} &
Method & Avg. & Fern & Flower & Fortress & Horns & Leaves & Orchids & Room & T-Rex \\
\cmidrule(lr){2-2} \cmidrule(lr){3-3} \cmidrule(lr){4-11}
&
NeRF~\cite{Mildenhall20} &  26.50 &  \bd{25.17} & 27.40 & 31.16 & 27.45 & 20.92 & 20.36 & 32.70 & 26.80  \\
&
\NeRFre &  26.66 &  24.93 & \bd{27.88} & 31.48 & 27.74 & \bd{21.12} & 20.35 & 32.77 & 27.04  \\
&
Transformer &  26.68 &  \underline{25.13} & \underline{27.87} & 31.42 & 27.79 & 21.10 & \bd{20.46} & 32.59 & 27.05  \\
&
Pool &  \bd{26.79} &  25.07 & 27.81 & \underline{31.50} & \underline{27.82} & \underline{21.11} & \underline{20.40} & \bd{33.14} & \bd{27.49}  \\
&
MLPMix &  \underline{26.76} &  25.01 & 27.85 & \bd{31.51} & \bd{27.88} & 21.09 & 20.38 & \underline{32.93} & \underline{27.45}  \\
\bottomrule

\end{tabular}
\label{tab:res:arch}
\end{table}
}

\newcommand{\tabSota}{
\begin{table}[t]
\centering
\small
\caption{{\bf Comparison with the state-of-the-art.}
\capNerf
Highlighted are the \bd{best} and \underline{second best} results.}
\hspace*{-1cm}
\begin{tabular}{r@{~~}c@{~~}c@{~~}c@{~~}c@{~~}c@{~~}c@{~~}c@{~~}c@{~~}c} \toprule
& SRN~\cite{Sitzmann19} & NV~\cite{Lombardi19} & LLFF~\cite{Mildenhall19} & NeRF~\cite{Mildenhall20} & \NeRFre & NSVF~\cite{Liu20} & IBRNet~\cite{Wang21} & GRF~\cite{Trevithick20} & \NeRFour \\ %
\cmidrule(lr){1-9} \cmidrule(lr){10-10}
\emph{Blender} PSNR & 22.25 & 26.05 & 24.88 & 31.01 & 31.83 & 31.75 & 28.14 & \underline{32.06} & \bd{32.34} \\
\emph{Blender} SSIM & 0.846 & 0.893 & 0.911 & 0.947 & 0.954 & 0.954 & 0.942 & \bd{0.960} & \underline{0.957} \\
\emph{LLFF-NeRF} PSNR & 22.84 & / & 24.13 & 26.50 & 26.66 & / & \underline{26.73} & 26.64 & \bd{26.76} \\
\emph{LLFF-NeRF} SSIM & 0.668 & / & 0.798 & 0.811 & 0.820 & / & \bd{0.851} & \underline{0.837} & 0.822 \\
\bottomrule
\end{tabular}
\label{tab:res:sota}
\end{table}
}

\begin{document}

\maketitle

\begin{abstract}
Neural radiance fields (NeRF) methods have demonstrated impressive
novel view synthesis performance. The core approach is to
render individual rays by querying a neural network at points sampled along
the ray to obtain the density and colour of the sampled points,
and integrating this information using the rendering equation.
Since dense sampling is computationally prohibitive, a common solution is to
perform coarse-to-fine sampling.

In this work we address a clear limitation of the vanilla coarse-to-fine
approach -- that it is based on a heuristic and not trained end-to-end
for the task at hand.
We introduce a differentiable module that learns to propose samples and their importance for the fine network,
and consider and compare multiple alternatives for its neural architecture.
Training the proposal module from scratch can be unstable due to
lack of supervision, so an effective pre-training strategy is also put forward.
The approach, named `NeRF in detail' (\NeRFour), achieves superior view synthesis quality
over NeRF and the state-of-the-art
on the synthetic Blender benchmark and on par or better performance on the
real LLFF-NeRF scenes.
Furthermore, by leveraging the predicted sample importance,  
a 25\% saving in computation can be achieved without
significantly sacrificing the rendering quality.
\end{abstract}

\section{Introduction}

We address the classic problem of view synthesis, where given multiple images
of a scene taken with known cameras, the task is to faithfully generate
novel views as seen by cameras arbitrarily placed in the scene.

In particular, we build on top of NeRF~\cite{Mildenhall20},
the recent impressive approach that represents a scene with a neural network;
where, for each pixel in an image, points are sampled along
the ray connecting the camera centre and the pixel,
the network is queried to produce color and density estimates,
and this information is integrated to produce the pixel color.
The samples are obtained through a heuristic hierarchical coarse-to-fine
approach that is not trained end-to-end.
The method is reviewed in Section~\ref{sec:nerfoverview}.

Our main contribution is the introduction of a `proposer' module
which makes the coarse-to-fine sampling procedure differentiable
and amenable to learning via gradient descent.
This enables us to train the entire network jointly for the end task.
However, training it from scratch is challenging as lack of supervision
causes instabilities and produces inferior results.
Therefore, we also propose an effective two-stage training strategy
where the network is first trained to mimic vanilla NeRF,
and then made free to learn better sampling strategies.
This approach, named `NeRF in detail' (\NeRFour), yields better scene representations,
achieving state-of-the-art results on two challenging view
synthesis benchmarks.

We consider a range of architectures for the `proposer' module.
Furthermore, the `proposer' can be trained to produce importance estimates
for the sample proposals, which in turn enables us to adaptively filter out
the least promising samples and reduce the amount of computation
needed to render a scene, without compromising on the rendering quality.

\subsection{Related work}

\paragraph{View synthesis.}
Classical approaches directly interpolate the
known images~\cite{Chen93,McMillan95,Seitz96,Debevec96,Fitzgibbon03a,Chaurasia13}
or the light field~\cite{Levoy96,Gortler96}, without
requiring the knowledge of the underlying scene geometry, but they often
need dense sampling of the scene in order to work well.
More recently, learning based approaches were used to blend
the images~\cite{Flynn16,Hedman18,Wang21a},
or to construct a multiplane image representation of the scene
which can be used to synthesize novel views~\cite{Zhou18,Mildenhall19,Li20a}.
Another set of methods generates the views by querying an explicit
3D representation of the scene,
such as a mesh~\cite{Koch95,Debevec96b,Shan13,Riegler20},
a point cloud~\cite{Grossman98,Zwicker01,Bui18,Aliev20},
or a voxel grid~\cite{Seitz97,Kutulakos98,Penner17},
but forming this 3D representation can be fragile.
We follow an increasingly popular approach
where scene geometry and appearance are represented implicitly using
a neural network~\cite{Park19,Mescheder19,Sitzmann19,Niemeyer20,Genova20,Mildenhall20}
and views are rendered by ray tracing;
NeRF~\cite{Mildenhall20} is reviewed in Section~\ref{sec:nerfoverview}.

\paragraph{Proposals.}
A popular approach in object detection is to first generate a number of
proposals, aiming to cover the correct object with high recall,
and then classification and refinement is applied to improve the
precision~\cite{Sande11,Girshick14,Girshick15,Ren15}.
Two-stage temporal action detection approaches~\cite{Shou16,Xu17,Zhao17}  %
follow a similar strategy, where the proposals are temporal segments,
and are therefore 1D as in our work where the proposals correspond to the distance
from the camera along the ray.
However, in both domains full supervision on the locations is available to train the proposers,
whereas this level of supervision is not available to us.
Differentiable resampling for particle filters~\cite{Jonschkowski18,Zhu20}
can also be seen as a coarse-to-fine approach akin to our proposer,
but there the sampling is based purely on coarse samples' locations and weights,
while our proposer also makes use of features computed by the coarse network.
\subsection{Overview of \emph{Neural radiance field} (NeRF)}
\label{sec:nerfoverview}

Here we give a brief overview of NeRF with a particular focus on
the issues relevant to this work; full details are available in the original
paper~\cite{Mildenhall20}.

NeRF models the radiance field with a neural network --
when queried with 3D world coordinates and 2D viewing direction,
the network outputs the density of the space, $\sigma$, and the view-dependent color, $\mathbf{c}$.
Rendering an image seen from any given camera can be done by shooting
a ray through each pixel and computing its color.
The ray color is computed independently for each ray, by sampling points
along the ray, querying the network for the points' densities and colors,
and integrating the information from all the samples via
the rendering equation:
\begin{equation}
\mathbf{\hat{C}} = \sum_{i=1}^N w_i \mathbf{c_i},~
w_i = T_i \left(1-\exp\left(-\sigma_i (t_{i+1}-t_i)\right)\right),~
T_i = \exp\left(-\sum_{j=1}^{i-1} \sigma_j (t_{i+1}-t_i)\right)
\label{eq:render}
\end{equation}
\noindent where $\sigma_i$, $\mathbf{c_i}$, $t_i$
is the $i$-th sample's density, color and location along the ray, and
$w_i$ is its contribution to the ray color estimate $\mathbf{\hat{C}}$.

The network is trained by sampling rays from all pixels of
the training set images,
and minimizing the L2 loss between the predicted and ground truth
ray color.
All the operations mentioned so far are differentiable and the network
is trained with gradient descent.

With an appropriate choice of the network architecture~\cite{Zhang20},
the density does not depend on the viewing direction,
and the color is somewhat restricted in how much it can vary
across viewpoints (a trade-off between not overfitting to training views
and modelling non-Lambertian effects).
This enables NeRF to produce consistent renderings across different
viewing directions.

\figNerfGen

One underlying assumption is that accurate ray rendering can be achieved
with a finite number of samples along the ray. With a larger number of
samples, the rendering should get better
but becomes computationally prohibitive.
This is why NeRF follows a coarse-to-fine approach, as illustrated in
Figure~\ref{fig:nerfgen}(a).
Namely, two networks -- coarse and fine -- are used,
where
(i) the coarse network is queried on a few, $N_c$, equally spaced samples
along the ray,
(ii) its outputs are used to obtain more, $N_f$, samples, and
(iii) the fine network is queried on the union of the samples
($N_c+N_f$) and produces the final rendering.
The sampling of $N_f$ points is done from a piece-wise constant probability density function,
where the pdf is computed via a handcrafted procedure involving the
densities produced by the coarse network (Figure~\ref{fig:nerfgen}(b)).
In brief, the rendering equation can be used on the coarse network's outputs
to calculate the contribution (\ie the weight, $w_i$ in eq.~\eqref{eq:render})
of every coarse sample to the final color,
and these weights are normalized to define the piece-wise constant pdf.
The intuition behind this heuristic is that the region of space
that contains the object closest to the camera
(\ie contributing most to the rendered color) should be sampled
the most in order to reveal details captured by the fine network.
The whole coarse-to-fine system is not trainable end-to-end
but the coarse network is trained independently with the same
reconstruction loss.

\paragraph{Recent developments.}
Due to its impressive performance, NeRF has attracted a lot of attention
in the field and has been extended in a variety of ways.
Many works adapted NeRF to situations it does not handle out of the box,
such as
real world scenes with varying lighting and transient objects~\cite{Martinbrualla20},
deformable scenes~\cite{Pumarola2021,Park20},
video~\cite{Li20,Xian21},
unknown cameras~\cite{Wang21}, \etc.
NeRF requires retraining for every scene and a few approaches
alleviate this requirement by sharing parameters across
scenes~\cite{Kosiorek21,Yu21a,Trevithick20}.
Combining NeRF with GANs yields promising 3-D aware image
generation~\cite{Chan21,Schwarz20}.
However, not many works have focused on improving the `core'
algorithm behind NeRF, with the exception of
NeRF++~\cite{Zhang20} which proposes a better parametrization
for large-scale unbounded real-world scenes,
and mip-NeRF~\cite{Barron21} which addresses aliasing issues.
In this work we improve a core component of NeRF that is the
hierarchical coarse-to-fine sampling of points along rays,
by replacing the heuristic non-trainable sample proposal module
with a fully end-to-end trainable one.
This is complementary to many of the above-mentioned approaches,
\eg~\cite{Martinbrualla20,Park20,Xian21,Barron21,Wang21a,Kosiorek21,Garbin21,Trevithick20},
since they also use coarse-to-fine sampling,
and our module can easily be swapped in.

\section{Learning to sample}

In this section we describe an improved coarse-to-fine approach in order
to facilitate better rendering of details.
As explained in the previous section, the coarse-to-fine procedure
used in the original NeRF work is a handcrafted heuristic
-- it matches intuition but it is likely suboptimal.
The coarse network is trained independently for reconstruction,
and it is not even `aware' of the fine network
(no error signal from the fine network reaches the coarse network).
Therefore, the coarse network is unable to adjust its outputs to benefit
the real end-goal of interest --
the reconstruction quality as produced by the fine network.

We replace this non-trainable sample proposal procedure
with a differentiable module (Figure~\ref{fig:nerfgen}(c))
and train the whole system end-to-end.
As before, the ray is sampled at regular intervals and the outputs
of the coarse network
are used to decide where to query the fine network.
Instead of just using the densities and a hand-engineered proposal
mechanism, as in the original NeRF,
we pass the features produced by the coarse network at each coarse sample
(the features are the values of the last activations before the projection
that produced the density) through a neural network that directly
produces the locations of the fine samples.
So, the input to the ``proposer'' is a sequence of $N_c$ coarse-network-produced
feature vectors and their positions along the ray
(scalars normalized to $[0, 1]$ such that $0$ corresponds to the near
and $1$ to the far plane),
and the output is a new set of $N_f$ positions along the ray
(again in $[0, 1]$ obtained by passing the scalar output through a sigmoid)
where the fine network will be queried.

\figProposer

\paragraph{Architectures.}
We consider a few options for proposer architecture
(Figure~\ref{fig:proposer}).
Here we outline the main features of the architectures and provide
full details in \isArXiv{Appendix~\ref{sec:app:arch}}{the supplementary material}.
The core design choice is to make the proposer network small relative
to the coarse and fine networks, in order to have negligible
additional parameters and computational overhead.

\emph{1.\ Transformer}:
A transformer~\cite{Vaswani17}
encoder processes the input features, and a transformer decoder
produces the final set of proposals
via cross attention with $N_f$ learnt queries.

\emph{2.\ Pool}:
Each point feature is summed up with the positional encoding of its
location along the ray, and passed through a fully connected layer (FC)
followed by a ReLU non-linearity. These processed features are
averaged together to form a single ``ray representation'' vector.
This vector is then passed through an FC to produce
the final set of proposals, where the output dimension of the FC is $N_f$.
The motivation behind the first stage is that simply average pooling
(feature+positional encoding) vectors would lose the feature-position
association due to all operations being linear,
so a non-linearity is required before the pooling.
The validity of this choice is confirmed by the ablations in Section~\ref{sec:results}.

\emph{3.\ MLPMix}:
The ``ray representation'' is obtained by applying a single
MLPMix~\cite{Tolstikhin21} block to the stacked point features
sorted by their position along the ray, followed by average pooling.
The final set of proposals are again decoded via an FC.

\subsection{Effective training}
\label{sec:efftrain}

As will be shown in Section~\ref{sec:results}, simply training the entire network
(coarse + proposer + fine networks) jointly from scratch
sometimes works, but sometimes fails to reach good performance.
We hypothesize that this is because the supervision is too weak
(just the final color of the rendered ray).
There is a chicken and egg problem --
training the proposer requires a good fine network that will provide
a useful training signal,
but the fine network can only become good if it is sampled at informative
locations.
It can be hard to stop this vicious cycle when training from scratch.

We note that the ``proposer'' idea has some resemblance to
two-stage object detectors such as Faster R-CNN~\cite{Ren15},
where the first network produces object proposals
and the second network examines the proposals in more detail.
The analogy with object detection and in particular DETR~\cite{Carion20}
inspired our Transformer proposer.
However, in object detection the proposal network is typically trained
with full supervision using object bounding box annotations,
while such supervision is not available in our case.

To alleviate the training difficulties, we divide training into two stages.
In the \emph{first stage}, the coarse and fine networks are trained
as in the vanilla NeRF, \ie using the heuristic proposer.
The trainable proposer is trained to predict the proposals
coming from the heuristic proposer --
the two sets of proposals are matched in a greedy fashion
(for speed reasons) and the sum of the L2 distances between
the matching proposals is minimized.
At the end of the first stage, the coarse and fine networks
are identical to the vanilla-NeRF ones,
while the proposer hopefully mimics the heuristic proposer.
In the \emph{second stage}, we swap in the learnt proposer
(\ie the samples seen by the fine network come from the learnt
proposer instead of the heuristic one)
and simply train the whole system end-to-end to minimize
the reconstruction loss of the fine network.
To prevent the system from diverging, we still add
the reconstruction loss of the coarse network as well,
but the coarse network is now optimized to both
minimize its reconstruction loss as well as produce
good features that enable the proposer to provide
informative samples to the fine network and yield good
fine network reconstructions.

The two training stages are equal in duration (number of SGD steps)
and for fair comparison with vanilla NeRF the total training time
is kept constant.

\subsection{Learning sample importance}
\label{sec:importance}

It is also possible to predict the ``importance'' of each proposal sample.
This can be useful to reduce the computational burden of ray rendering
as the fine network can be queried only with samples deemed to be important.
The proposer can predict importance scores for all the samples
by simply regressing another set of $N_c+N_f$ scalars
using the same mechanism as with the proposal generation,
\eg for the \emph{Pool} and \emph{MLPMix} architectures
another FC head operating on the ``ray representation'' vector
produces the $N_c+N_f$ scalars.

To make another parallel with two-stage object detectors --
typically the region proposals are assigned confidence scores,
facilitated by the fully supervised training.
Here, unlike in object detection, such fine-grained supervision is not available,
but the output of the fine network can be used as the training signal instead.
Namely, the weight $w_i$ (eq.~\eqref{eq:render})
that each fine sample contributes to the final rendering
is exactly what the importance prediction aims to forecast
-- iff it is high then the sample is important.
So the weights are thresholded at the desired accuracy level
(0.03 is used as a reasonable value, we did not investigate others)
to define the positive and negative samples,
and the importance predictor is trained to predict them
using the balanced logistic regression loss.

\section{Experiments and discussion}

In this section we evaluate the performance of our `NeRF in detail' (\NeRFour) method.
First, the experimental protocol and the benchmarks are described,
followed by the comparison of various proposer architectures.
Next, our method with trainable proposer
is contrasted against NeRF and the state-of-the-art in terms of view synthesis quality and speed.
Finally, we discuss relation to other methods, limitations
and potential avenues for future research.

\subsection{Experimental protocol, datasets and evaluation}
\label{sec:res:protocol}
\isArXiv{}{\vspace{-0.2cm}}

\paragraph{Datasets.}
The main benchmarks from the original NeRF~\cite{Mildenhall20} are used,
namely \emph{Blender}~\cite{Mildenhall20} and \emph{LLFF-NeRF}~\cite{Mildenhall19,Mildenhall20}.
\emph{Blender} is a realistically rendered $360^{\circ}$ synthetic dataset comprising of 8 scenes
with 100 training, 100 validation and 200 testing views
of resolution $800 \times 800$.
\emph{LLFF-NeRF} contains forward-facing $1008 \times 756$ images
of 8 real scenes,
ranging between 20 and 62 images per scene,
$1/8$ of which are used for testing and the rest for training;
since there is no validation set we use the training set in its place.

\paragraph{Performance metrics.}
Following standard procedure~\cite{Mildenhall20},
novel view synthesis quality is measured using
peak signal-to-noise ratio (PSNR),
and structural similarity (SSIM)~\cite{Wang04},
where higher scores signify better performance.
We observe very similar trends for the two metrics, so \isArXiv{here we mainly report}{the main paper mainly
reports} PSNR and the SSIM for all experiments can be found in
\isArXiv{Appendix~\ref{sec:app:results}}{the supplementary}.
For most important experiments we train five times with different
random seeds.

\paragraph{Training procedure.}
We use the same optimization procedure and hyper-parameters
for all experiments, datasets and scenes.
As with the original NeRF, $N_c=64$ points per ray are processed by the coarse network,
while the fine network evaluates 192 samples
(reusing the same $N_c=64$ locations as the coarse network and additional $N_f=128$ samples
obtained from the sample proposal procedure).
Batches are compiled by sampling rays randomly across the entire training set,
and optimization is ran for 10 billion rays with a
total batch size of 66k (\ie roughly 150k SGD steps).
The Adam optimizer~\cite{Kingma15} (with default hyper-parameters) is used with
the learning rate first following a linear warmup from 0 to $5 \times 10^{-4}$
for 1k steps and then switching to the cosine decay schedule.
The checkpoint with the best PSNR on a random subset of the validation set is
used (\ie early stopping),
although in the vast majority of scenes this corresponds to the last checkpoint.
The entire system is implemented in JAX~\cite{JAX18}
\isArXiv{and the DeepMind JAX Ecosystem~\cite{deepmind2020jax}}{}
and trained on 16 Cloud TPUs for about 10 hours.
\label{checklist:compute}

Note that our training procedure is slightly different from the original NeRF,
\eg they train with smaller batch sizes and for fewer iterations
(around 1.2 billion rays).
We also make sure that each ray passes through the center of its
pixel instead of the corner.
Our reimplementation, which we name \NeRFre, outperforms the original and
thus serves as a good baseline for our \NeRFour.

\setcounter{topnumber}{5}
\setcounter{totalnumber}{5}
\renewcommand{\topfraction}{0.9}
\tabArch
\tabSota

\subsection{Results and discussion}
\label{sec:results}
\isArXiv{}{\vspace{-0.2cm}}

\paragraph{Proposer architectures.}
The performance of the three proposer architectures
-- Transformer, Pool and MLPMix --
is compared in Table~\ref{tab:res:arch}.
While all three often work well and there are scenes on which
each of them performs best, the Transformer is inferior
due to significantly worse performance on a few scenes
(\eg \emph{Blender: Ficus}, \emph{Blender: Lego}, \emph{LLFF-NeRF: T-Rex})
and we observe its training curves are somewhat unstable.
Pool and MLPMix are somewhat on par, but Pool is significantly worse
on two scenes (\emph{Blender: Materials}, \emph{Blender: Mic}) and overall
has a higher variance.
Therefore, in the rest of the paper, our method \NeRFour uses
the MLPMix proposer.
\isArXiv{Appendix~\ref{sec:app:results}}{The supplementary material} contains additional ablations on the
proposer architectures, including
various ways to integrate sample positions into the Pool architecture,
and a ``blind'' proposer
(proposer which ignores input features and therefore always outputs
the same proposals; its bad performance verifies our good proposers
are not exploiting some aspect of the datasets via a simple
degenerate strategy).

\paragraph{Two-stage training.}
Training from scratch often succeeds but in the majority of those
cases it underperforms our two-stage training procedure.
It also often fails badly
(\eg $-5$ PSNR on \emph{Blender: Ship}), validating the need for
the two-stage training.
Full results are available in \isArXiv{Appendix~\ref{sec:app:results}}{the supplementary}.

\paragraph{Comparison with the state-of-the-art (Tables~\ref{tab:res:arch} and~\ref{tab:res:sota}).}
The MLPMix-based proposer consistently outperforms NeRF and \NeRFre
on all \emph{Blender} scenes, and is on par or better on the \emph{LLFF-NeRF}
scenes, verifying the effectiveness of our approach;
the improvements are statistically significant,
as demonstrated in \isArXiv{Appendix~\ref{sec:app:results}}{the supplementary material}.
Particularly impressive is its performance on
\emph{Ficus} and \emph{Lego} in \emph{Blender},
achieving $+1.84$ and $+1.17$ PSNR ($-53\%$ and $-31\%$ MSE), respectively,
and $+0.41$ on \emph{LLFF-NeRF: T-Rex} ($-10\%$ MSE).
Furthermore, \NeRFour outperforms state-of-the-art in PSNR
(the metric optimized by all approaches)
and is on par in SSIM.

\figSpeedupLlff
\figResQual

\paragraph{Qualitative results.}
Figure~\ref{fig:res:qual} contrasts synthesized views of our \NeRFour
versus \NeRFre. While \NeRFre shows impressive renderings,
\NeRFour does significantly better in capturing
some finer details, such as edges, thin structures and small objects.
These differences do not feature so prominently in the
quantitative evaluations since the proportion of the image that
contains such details is typically small,
but the improvements are clearly visible.

\paragraph{Speedup.}
Figure~\ref{fig:res:speedupllff} shows speedups obtained by
only querying the fine network on samples deemed to be important
by the proposer (Section~\ref{sec:importance}).
For most scenes in \emph{LLFF-NeRF} it is possible to render views
25\% faster without any loss in view synthesis quality.
\NeRFour consistently dominates \NeRFre,
producing better images using fewer computations.
For example, on the \emph{Fern} scene, \NeRFour that uses 27\% fewer
computations achieves better results than the \NeRFre,
while if \NeRFre uses 21\% fewer computations it
suffers a loss in PSNR of 0.8 ($+20\%$ MSE).
Speedups are even more dramatic on the \emph{Blender} dataset
(\isArXiv{Figure~\ref{fig:res:speedupblender}}{supplementary}) as many rays belong to the uniformly white background
and our method automatically learns to allocate them very
few samples.

Note that speeding up NeRF is not the main focus of this work,
and better approaches that are orders of magnitudes faster exist,
such as~\cite{Lindell21,Hedman21,Rebain21,Yu21,Reiser21}.
However,
(i) our approach essentially comes for free without heavily specialized
machinery,
(ii) it is compatible with some of the approaches
such as FastNeRF~\cite{Garbin21} and BakingNeRF~\cite{Hedman21}
since they still use coarse-to-fine sampling,
and
(iii)
some of them are not capable of handling
real world unbounded scenes~\cite{Reiser21,Yu21} like the ones in \emph{LLFF-NeRF}
or require significantly more storage to represent
a scene~\cite{Garbin21,Hedman21,Reiser21,Yu21}.

\paragraph{What is learnt?}
\isArXiv{
\newcommand{\whatislearnt}{
\figWhatLearnt
\afterpage{\FloatBarrier}

Figure~\ref{fig:whatlearnt} investigates what is being learnt.
The heuristic proposer of vanilla NeRF~\cite{Mildenhall20}
often provides good proposals, but it oversamples some regions
while sometimes undersampling the important surfaces.
Our learnt proposer rarely misses the important surfaces
and yields a much more diverse sampling.
This in turn yields a more accurate rendering,
but is also beneficial for effective training of the fine network.
Importance prediction
(\isArXiv{Section~\ref{sec:importance}}{Section~2.2})
does a good job at picking
the most important proposals (\ie proposals that are close to the
surface closest to the ray origin).
Furthermore, the proposals and their importance increase in density when the
(horizontal) ray is tangent or near to tangent to the surface.
}

\newcommand{\figWhatLearnt}{
\def\whatH{5cm}
\begin{figure}[t]
\centering
\hspace*{-1cm}
\begin{tabular}{cc}
\includegraphics[height=\whatH,valign=c]{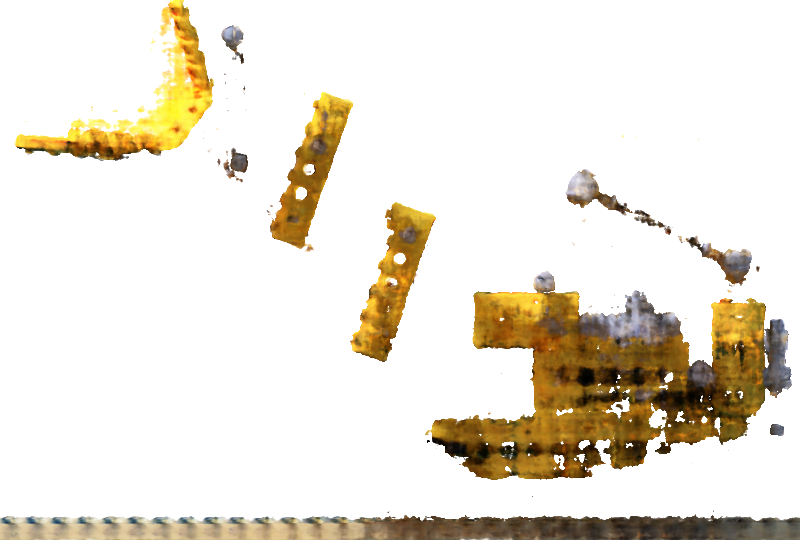}
&
\includegraphics[height=\whatH,valign=c]{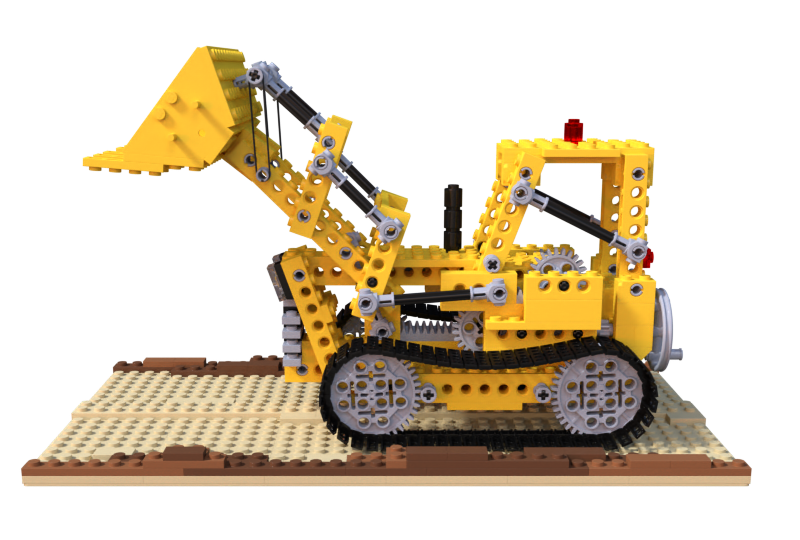}
\\ \addlinespace[0.3em]
(a) A cross section from the \emph{Blender: Lego} scene.
&
(b) Test image roughly aligned with (a)
\\ \addlinespace[1em]
\includegraphics[height=\whatH,valign=c]{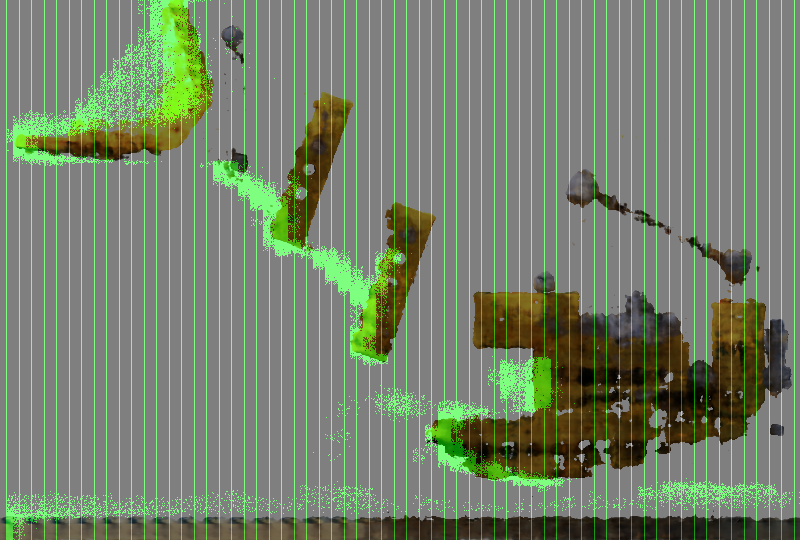}
&
\includegraphics[height=\whatH,valign=c]{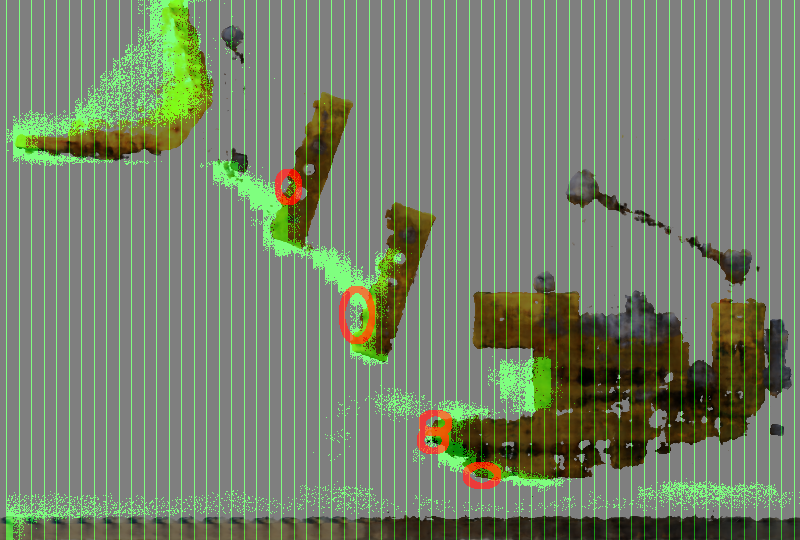}
\\ \addlinespace[0.3em]
(c) Heuristic proposals (+ coarse samples)~\cite{Mildenhall20}
&
(d) Highlighted mistakes in (c)
\\ \addlinespace[1em]
\includegraphics[height=\whatH,valign=c]{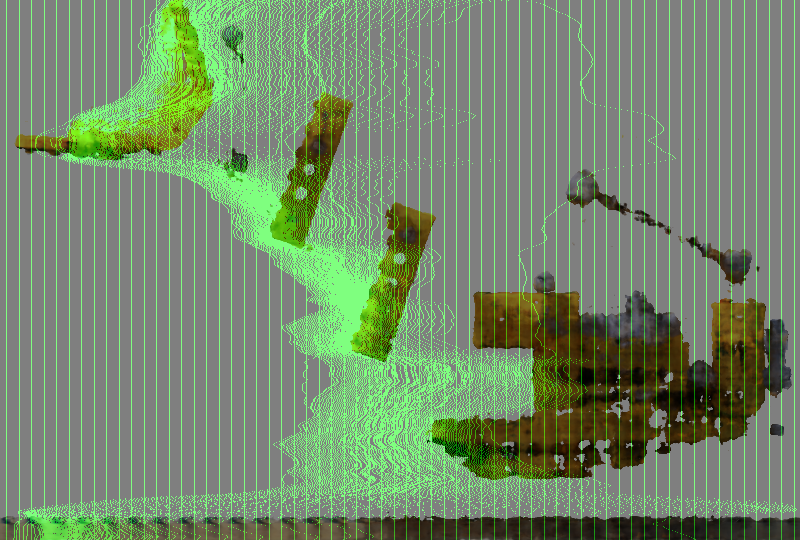}
&
\includegraphics[height=\whatH,valign=c]{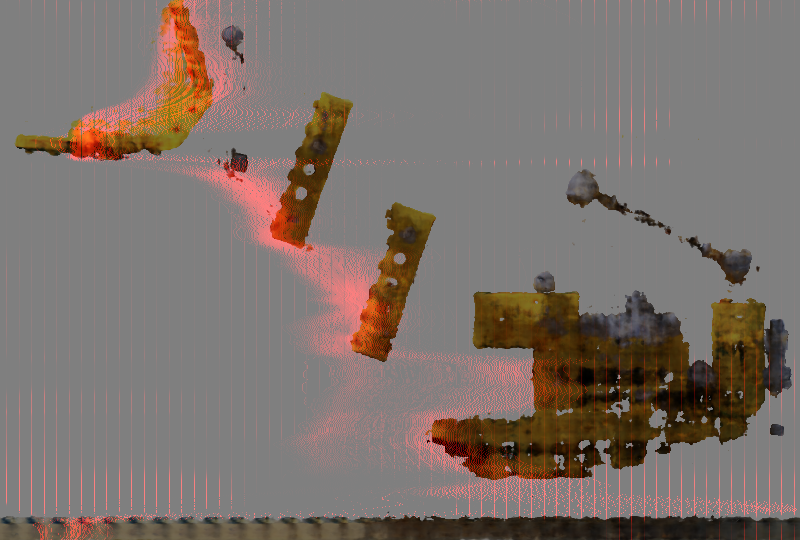}
\\ \addlinespace[0.3em]
(e) \NeRFour proposals (+ coarse samples)
&
(f) Importance prediction for \NeRFour proposals in (e)
\end{tabular}
\caption{{\bf What is learnt?}
(a) A cross section of the \emph{Blender: Lego} scene,
produced by querying the fine network densely on the plane
and plotting the color masked by the occupancy.
(b) An image from the test set whose camera plane is roughly
parallel to the cross section in (a).
(c) and (e) For each row of the cross section image,
a ray is shot from left to right and proposals (along with coarse samples)
are overlaid in green over the cross section image;
the samples are displayed at pixel resolution even though in reality they are real-valued.
(c) shows the heuristic proposals of our reimplementation \NeRFre of vanilla NeRF~\cite{Mildenhall20},
while (e) shows our \NeRFour learnt proposals.
(f) All proposals from (e) are colored in red such that the intensity
is proportional to the estimated importance (\isArXiv{Section~\ref{sec:importance}}{Section~2.2}).
The heuristic proposals are over-concentrated in a few areas and sometimes
undersample the surface closest to the ray origin (d).
\NeRFour proposals rarely miss the closest surface (e), and are much more diverse
which provides a more accurate rendering but also a better sampling
that facilitates training of the fine network.
Importance prediction does a good job at highlighting the
the most promising proposals (f).
}
\label{fig:whatlearnt}
\end{figure}
}

\whatislearnt
}{
We find that the proposer learns to densely sample the space near the surface
closest to the camera. This is an expected behaviour since these points
have the largest influence on the ray color.
Similarly, sample importance estimates tend to peak right at the
closest surface.
These behaviours are illustrated in the supplementary material.
}

\paragraph{Discussion, limitations and future work.} \label{sec:limits}
Mip-NeRF~\cite{Barron21} actually achieves
better PSNR performance on the \emph{Blender} dataset
but we do not include it in the comparison because
(i) it is a concurrent \isArXiv{}{non-peer-reviewed} approach,
(ii) it is not directly applicable on real unbounded scenes
like the ones comprising the \emph{LLFF-NeRF} dataset
(it achieves no improvements over NeRF),
and
(iii) its approach is complementary to ours.
In fact, \NeRFour can be applied on many works that extend NeRF
as long as they use or are amenable to using the coarse-to-fine
strategy when rendering rays.
This includes Mip-NeRF~\cite{Barron21},
both top competitors from Table~\ref{tab:res:sota}
(GRF~\cite{Trevithick20}, IBRNet~\cite{Wang21a}),
NeRFs for unconstrained~\cite{Martinbrualla20},
deformable~\cite{Park20} and temporal scenes~\cite{Xian21},
as well as some approaches for speeding up NeRF~\cite{Hedman21,Garbin21}.

\NeRFour as well as other NeRF-based approaches still produce significant
errors especially on the real-world scenes, where on \emph{LLFF-NeRF}
the performances have somewhat plateaued
(\NeRFour , mip-NeRF~\cite{Barron21}, GRF~\cite{Trevithick20}, IBRNet~\cite{Wang21a}).
It is likely that the bottleneck is somewhere other than
the proposer mechanism,
even though we believe that our proposer module will be useful for future
NeRF-based approaches.
For example, deeper networks might be required to model these complex scenes,
but this in turns exacerbates NeRF's problem with being data hungry and
not utilizing the common structure of the world.
Few recent works have started tackling
the latter issue~\cite{Kosiorek21,Yu21a,Trevithick20}
and we believe the proposer has a place in these approaches as well --
learning a universal proposer shared across scenes can still facilitate learning
and focus the effort on relevant parts of the scene.
This work has kept the proposer very shallow with few parameters
in order to make \NeRFour directly comparable to NeRF,
but it is possible that a deeper proposer is required to
reach its full potential.

\isArXiv{\afterpage{\FloatBarrier}}{}
\section{Conclusions}
\isArXiv{}{\vspace{-0.3cm}}
We have presented a `proposer' module that
learns the hierarchical coarse-to-fine sampling, thus
enabling NeRF to be trained end-to-end for the view synthesis task.
Multiple architectures for the module are explored
and the best consistently outperforms the vanilla NeRF
on both challenging benchmarks.
\NeRFour fares well against the state-of-the-art with negligible overhead
in the number of parameters and amount of computation.
It also enables faster execution since 25\% of the samples can be
pre-filtered without sacrificing the view synthesis quality.

\isArXiv{
\paragraph{Acknowledgments.}
}{
\begin{ack}
}
We would like to thank Bojan Vujatovi\'c and Yusuf Aytar for helpful discussions.
\isArXiv{}{
\end{ack}
}

\isArXiv{}{\FloatBarrier}
{\small
\bibliographystyle{ieee}
\bibliography{bib/shortstrings,bib/more,bib/vgg_local,bib/vgg_other}
}

\isArXiv{}{
\section*{Checklist}

\begin{enumerate}

\item For all authors...
\begin{enumerate}
  \item Do the main claims made in the abstract and introduction accurately reflect the paper's contributions and scope?
    \answerYes{}
  \item Did you describe the limitations of your work?
    \answerYes{See Section~\ref{sec:limits}.}
  \item Did you discuss any potential negative societal impacts of your work?
    \answerNo{We don't see how improving view synthesis could have negative societal impacts.}
  \item Have you read the ethics review guidelines and ensured that your paper conforms to them?
    \answerYes{}
\end{enumerate}

\item If you are including theoretical results...
\begin{enumerate}
  \item Did you state the full set of assumptions of all theoretical results?
    \answerNA{}
	\item Did you include complete proofs of all theoretical results?
    \answerNA{}
\end{enumerate}

\item If you ran experiments...
\begin{enumerate}
  \item Did you include the code, data, and instructions needed to reproduce the main experimental results (either in the supplemental material or as a URL)?
    \answerNo{We provided a detailed description of the approach that is sufficient in order to reproduce it.}
  \item Did you specify all the training details (e.g., data splits, hyperparameters, how they were chosen)?
    \answerYes{See Section~\ref{sec:res:protocol} and the supplementary material.}
	\item Did you report error bars (e.g., with respect to the random seed after running experiments multiple times)?
    \answerYes{We repeated the most important experiments 5 times and report all results.
    Due to space constraints we only present selected results in the main paper, but we list
    all results of all experiments in the supplementary material.}
	\item Did you include the total amount of compute and the type of resources used (e.g., type of GPUs, internal cluster, or cloud provider)?
    \answerYes{We included the amount of computation and type of resources required each experiment in Section~\ref{checklist:compute}.}
\end{enumerate}

\item If you are using existing assets (e.g., code, data, models) or curating/releasing new assets...
\begin{enumerate}
  \item If your work uses existing assets, did you cite the creators?
    \answerYes{Section~\ref{sec:res:protocol}.}
  \item Did you mention the license of the assets?
    \answerNo{We are using existing assets.}
  \item Did you include any new assets either in the supplemental material or as a URL?
    \answerNo{}
  \item Did you discuss whether and how consent was obtained from people whose data you're using/curating?
    \answerNo{We don't think discussion is needed -- two datasets were used that were directly created by cited authors (rendered 3D scenes, and images captured by the authors).}
  \item Did you discuss whether the data you are using/curating contains personally identifiable information or offensive content?
    \answerNo{There is no personally identifiable information or offensive content in the data.}
\end{enumerate}

\item If you used crowdsourcing or conducted research with human subjects...
\begin{enumerate}
  \item Did you include the full text of instructions given to participants and screenshots, if applicable?
    \answerNA{}
  \item Did you describe any potential participant risks, with links to Institutional Review Board (IRB) approvals, if applicable?
    \answerNA{}
  \item Did you include the estimated hourly wage paid to participants and the total amount spent on participant compensation?
    \answerNA{}
\end{enumerate}

\end{enumerate}

}

\isArXiv{
\appendix
\ifdefined\isArXiv
\else

\documentclass{article}

\PassOptionsToPackage{numbers, sort}{natbib}
\usepackage{neurips_2021}

\usepackage[utf8]{inputenc} %
\usepackage[T1]{fontenc}    %
\usepackage{hyperref}       %
\usepackage{url}            %
\usepackage{booktabs}       %
\usepackage{amsfonts}       %
\usepackage{nicefrac}       %
\usepackage{microtype}      %
\usepackage{xcolor}         %

\usepackage{graphicx}
\usepackage{subcaption}
\usepackage{amsmath}
\usepackage{lipsum}

\usepackage{afterpage}
\usepackage{placeins}  %

\usepackage{multirow}
\usepackage[export]{adjustbox}  %

 \usepackage{import}  %

\usepackage{xspace}
\def\eg{\emph{e.g.}\@\xspace}
\def\ie{\emph{i.e.}\@\xspace}
\def\etal{\emph{et al.}\@\xspace}
\def\cf{\emph{c.f.}\@\xspace}
\def\etc{\emph{etc}}
\newcommand{\bd}[1]{{{\bf #1}}}
\newcommand{\st}[1]{{{\scriptsize #1}}}
\newcommand{\rotintab}[1]{\rotatebox[origin=c]{90}{#1}}

\renewcommand{\paragraph}[1]{\smallskip\noindent{\bf{#1}}}

\DeclareMathOperator*{\argmax}{arg\,max}
\newcommand{\loss}{\mathcal{L}}

\def\NeRFre{NeRF$^{\dagger}$\@\xspace}

\def\NeRFour{NeRF-ID\@\xspace}

\newcommand{\capNerf}{
NeRF denotes the results reported in the original paper~\cite{Mildenhall20},
\NeRFre is our reimplementation.
}

\newcommand{\isArXiv}[2]{#1}

\title{-- Supplementary material -- \\ NeRF in detail: Learning to sample for view synthesis}

\begin{document}

\maketitle

\tableofcontents

\vspace*{1cm}
\fi

\newcommand{\capScatter}{
For each scene and each method, the performance for all experimental runs
is plotted. Vertical black lines connect the 20\textsuperscript{th} and 80\textsuperscript{th} percentiles
(\ie the 2\textsuperscript{nd} best and 2\textsuperscript{nd} worst performance out of 5 runs),
indicating a robust estimate of the range of performances the method achieves.
The median performance is highlighted with its numerical value.
}

\newcommand{\vspaceScatter}{\vspace*{1cm}}

\newcommand{\figArchDetail}{
\begin{figure}[t]
\centering
  \begin{subfigure}[t]{\linewidth}
  \centering
  \includegraphics[width=\linewidth]{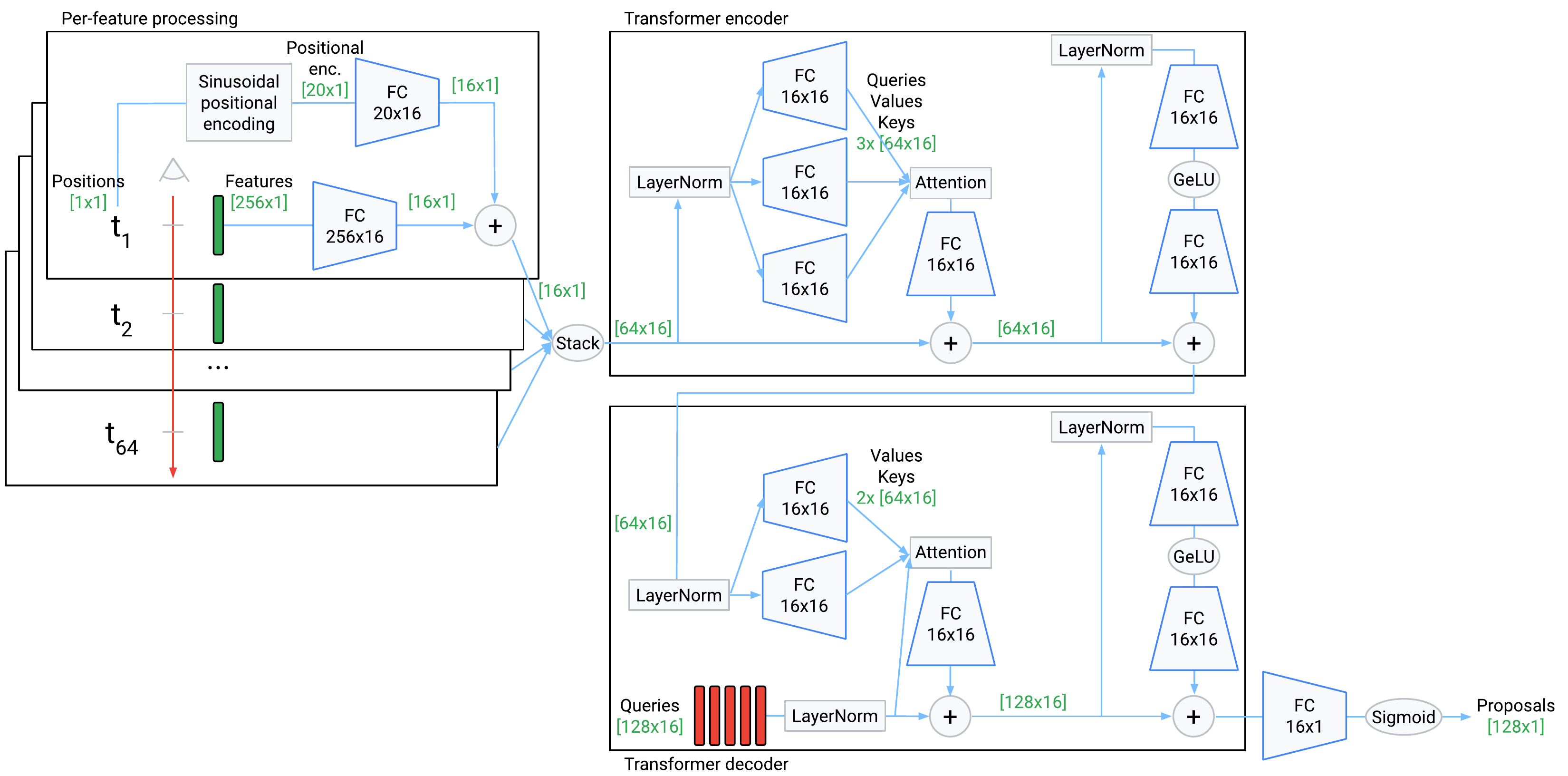}
  \caption{Transformer}
  \label{fig:archdetail:transf}
  \end{subfigure}
\\
  \begin{subfigure}[t]{\linewidth}
  \centering
  \includegraphics[width=\linewidth]{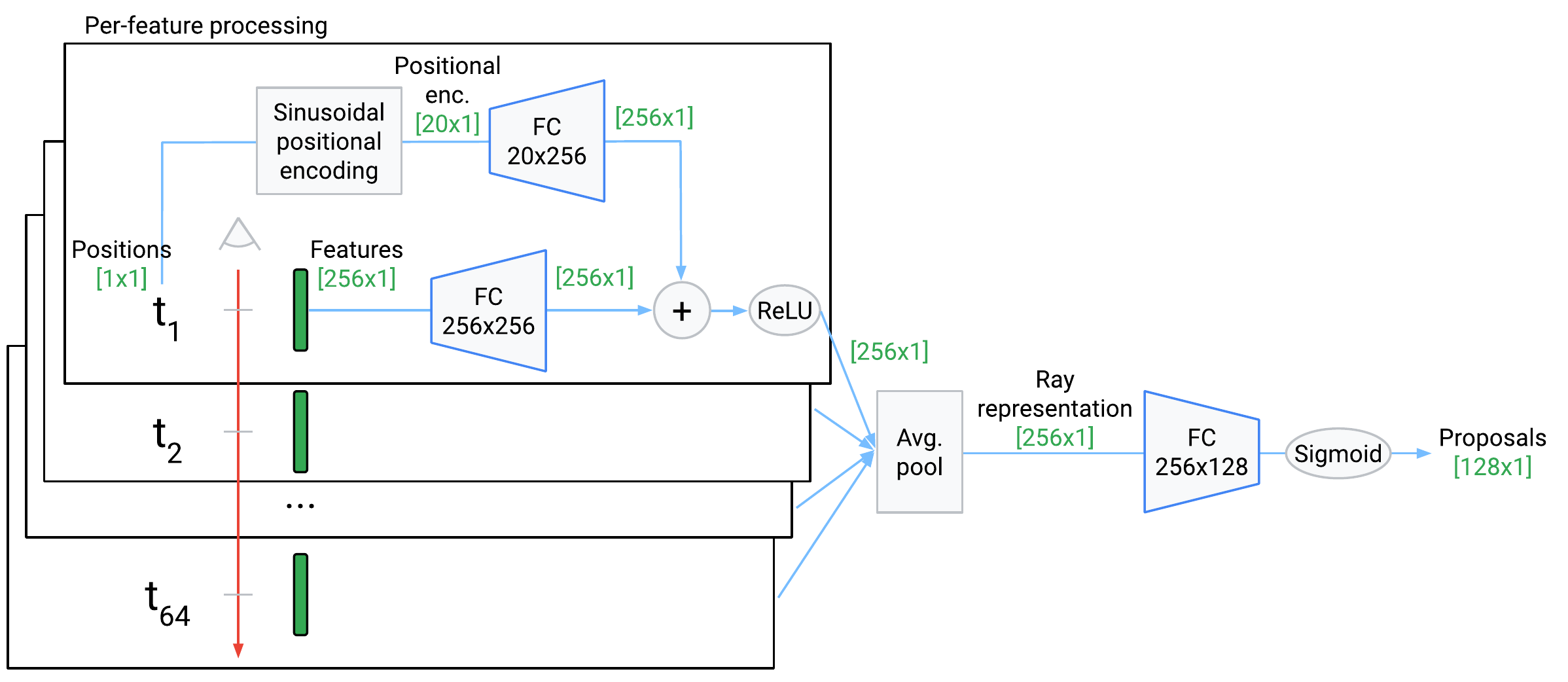}
  \caption{Pool}
  \label{fig:archdetail:pool}
  \end{subfigure}
\\
  \begin{subfigure}[t]{\linewidth}
  \centering
\includegraphics[width=\linewidth]{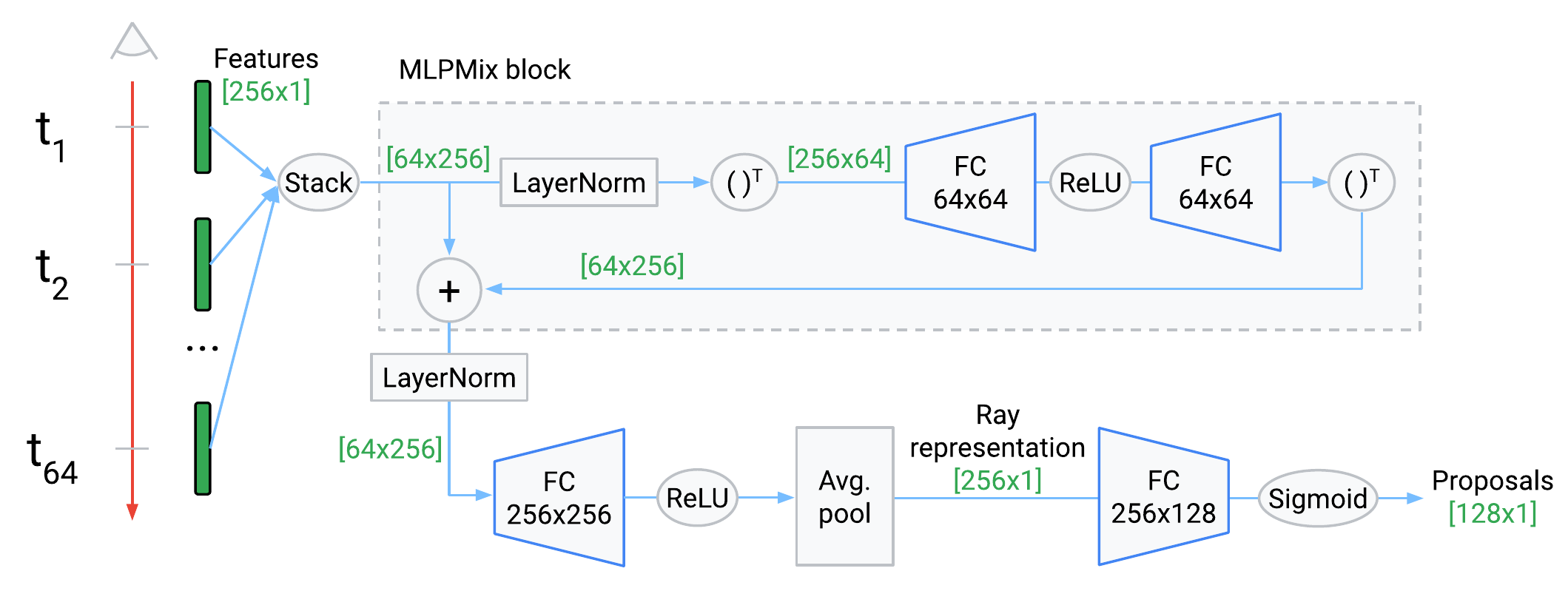}
  \caption{MLPMix}
  \label{fig:archdetail:mlpmix}
  \end{subfigure}
\caption{{\bf Trainable proposer architectures (detailed).}
Assumes the number of coarse samples is $N_c=64$ and the number of proposals is $N_f=128$,
as is default in our \NeRFour (same values are used in NeRF~\cite{Mildenhall20}).
}
\label{fig:archdetail}
\end{figure}
}

\newcommand{\figResArchPsnr}{
\begin{figure}[t]
\centering
\includegraphics[width=\linewidth]{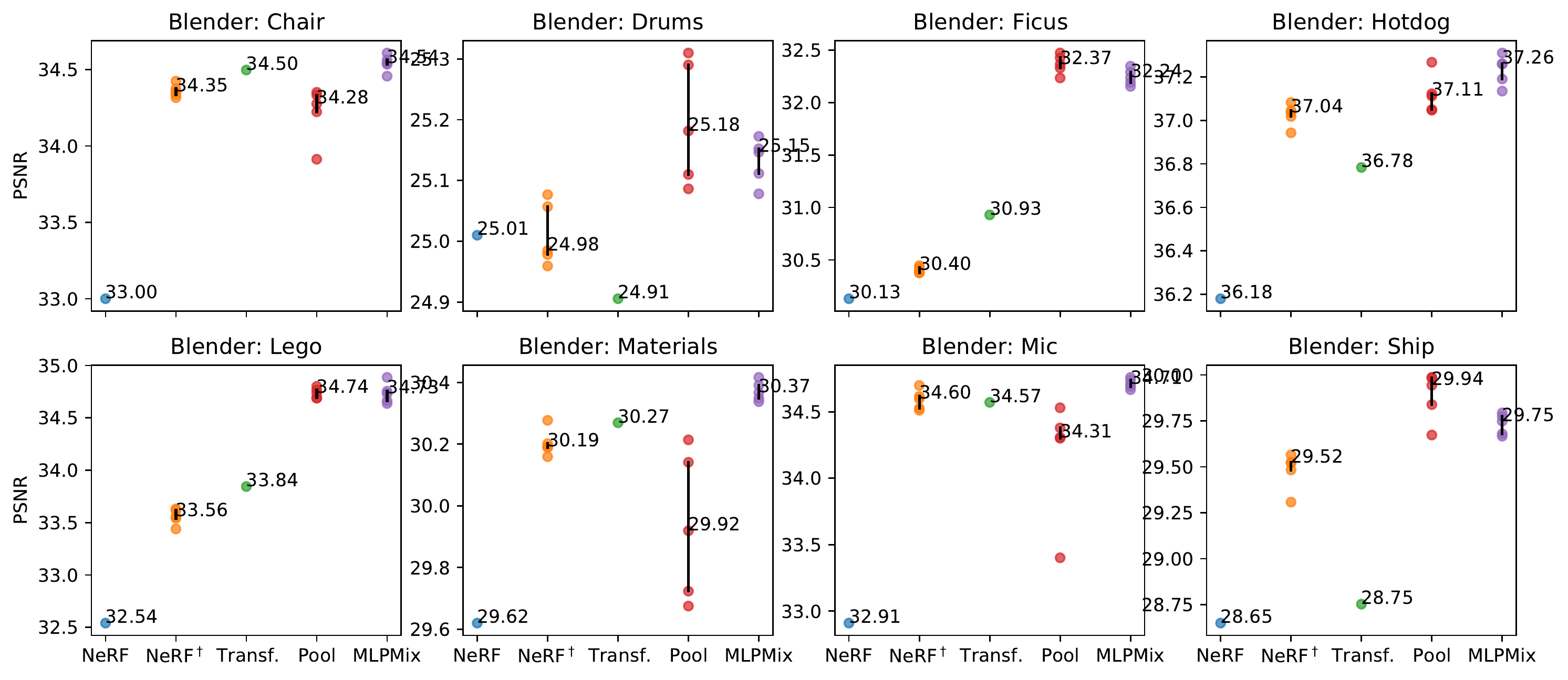} \\
\vspaceScatter
\includegraphics[width=\linewidth]{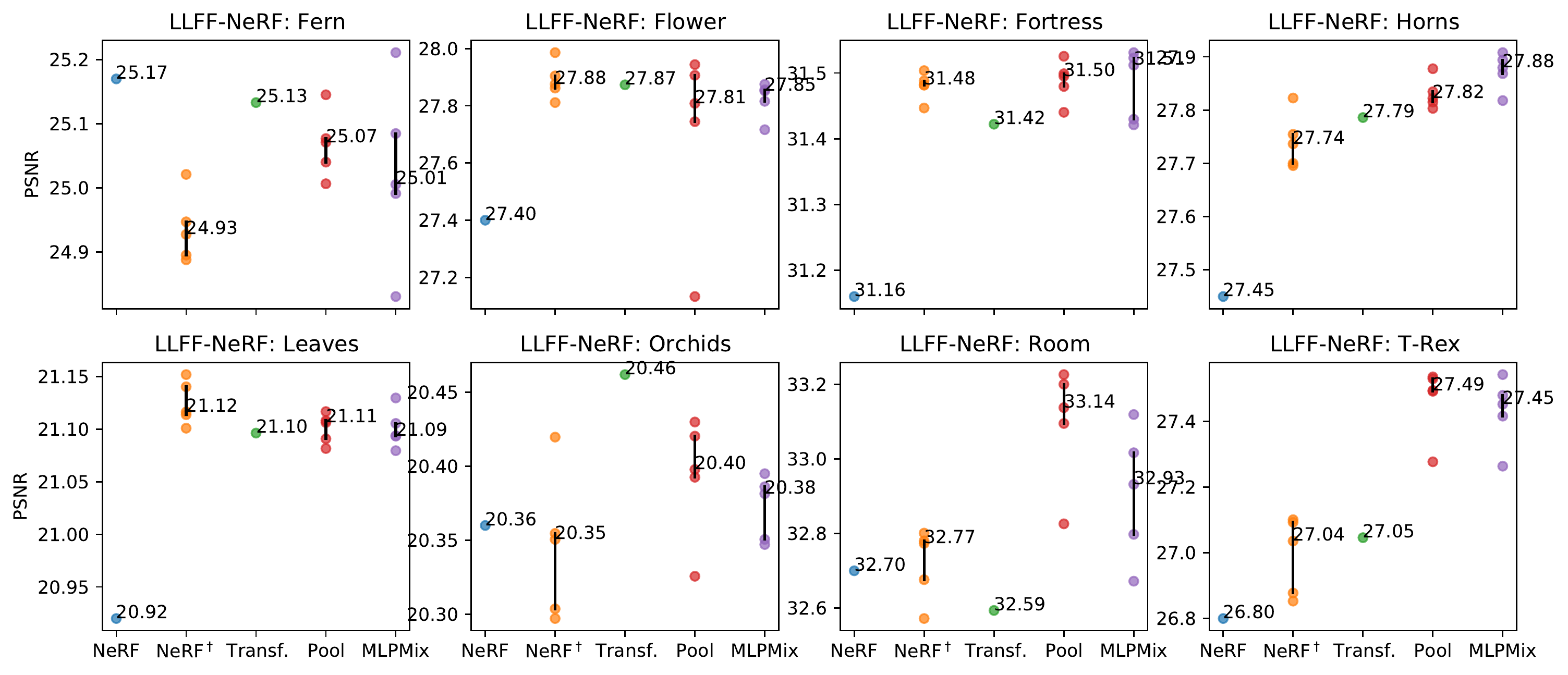}
\caption{{\bf Comparison of proposer architectures and vanilla NeRF (PSNR).}
\capScatter
\capNerf
}
\label{fig:res:arch:psnr}
\end{figure}
}

\newcommand{\figResArchSsim}{
\begin{figure}[t]
\centering
\includegraphics[width=\linewidth]{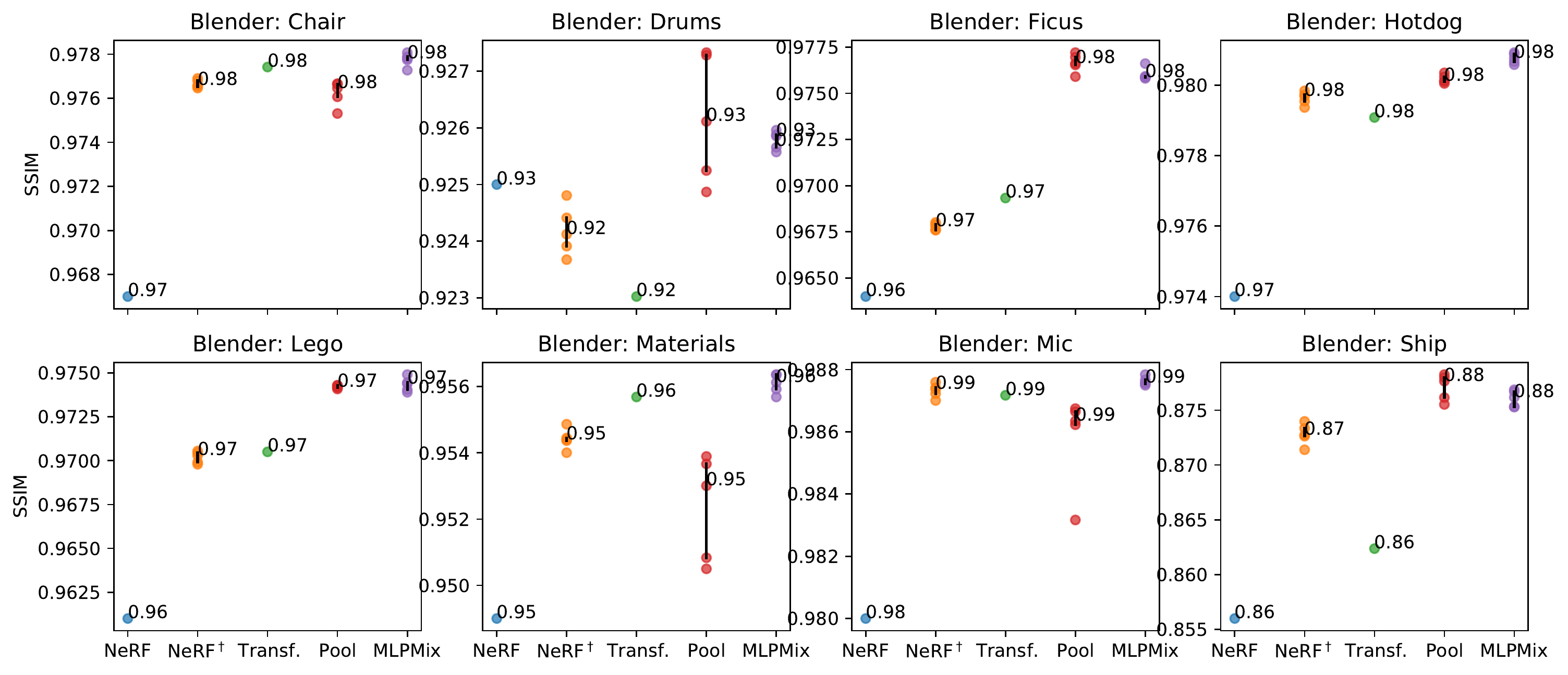} \\
\vspaceScatter
\includegraphics[width=\linewidth]{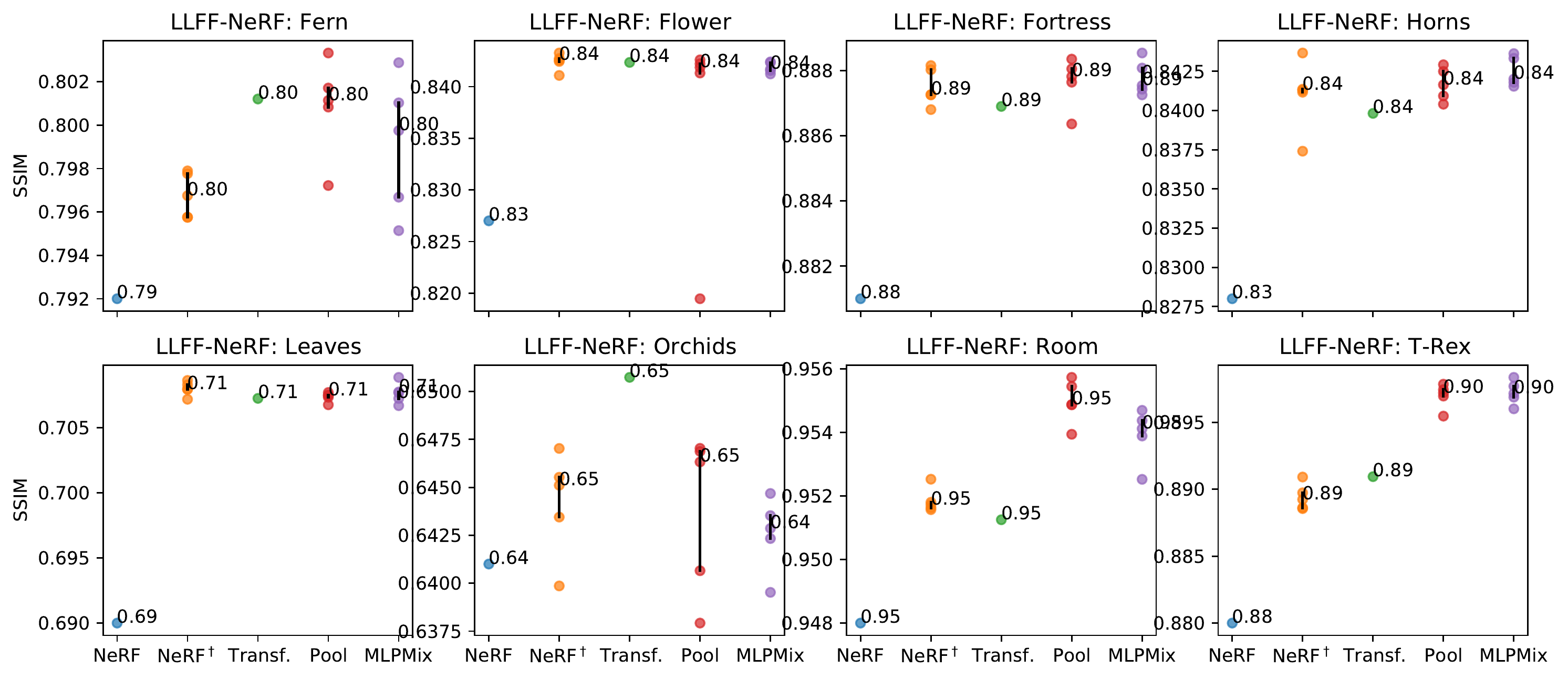}
\caption{{\bf Comparison of proposer architectures and vanilla NeRF (SSIM).}
\capScatter
\capNerf
}
\label{fig:res:arch:ssim}
\end{figure}
}

\newcommand{\figResMoreArchPsnr}{
\begin{figure}[t]
\centering
\includegraphics[width=\linewidth]{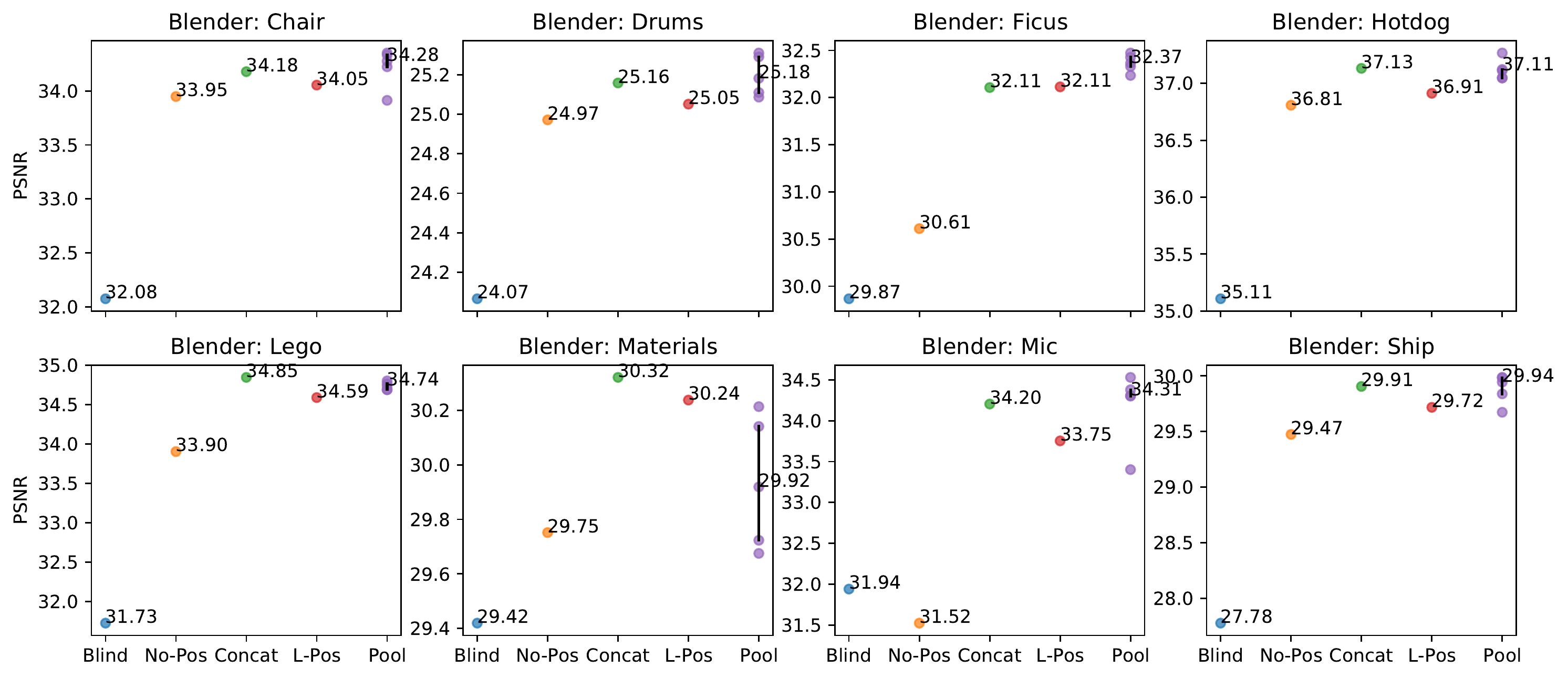} \\
\vspaceScatter
\includegraphics[width=\linewidth]{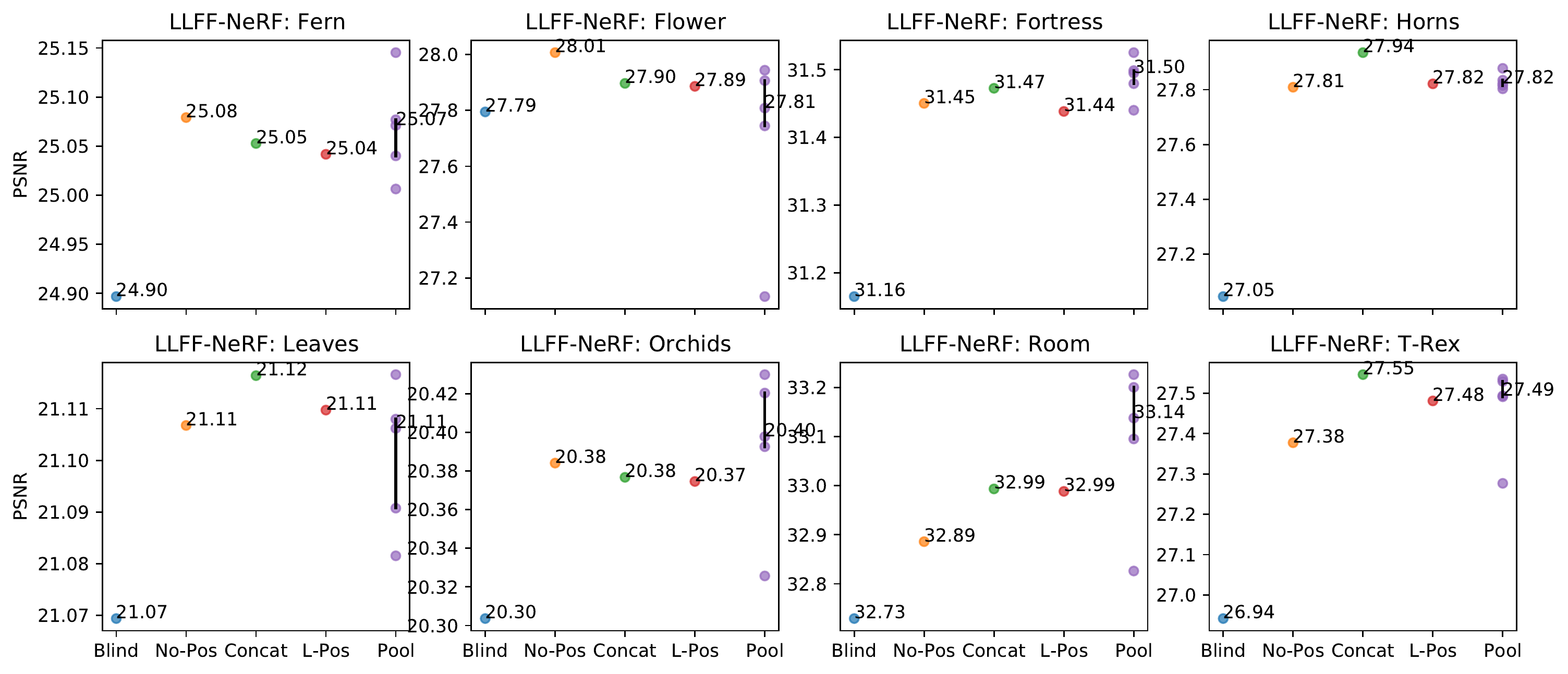}
\caption{{\bf Additional ablations on the \emph{Pool} proposer architecture.}
\capScatter
The \emph{Blind} proposer ignores input coarse-features
and therefore simply always outputs the same proposals.
The \emph{No-position} (No-Pos) proposer
has the same architecture as \emph{Pool} apart from
not using the positions of coarse samples along the ray.
\emph{Concat} concatenates the positional encodings instead of summing
them up as in \emph{Pool}.
\emph{Learnt-position} (L-Pos) learns positional encodings instead
of using the sinusoidal positional encodings as in \emph{Pool}.
}
\label{fig:res:morearch:psnr}
\end{figure}
}

\newcommand{\figResScratch}{
\begin{figure}[t]
\centering
\includegraphics[width=\linewidth]{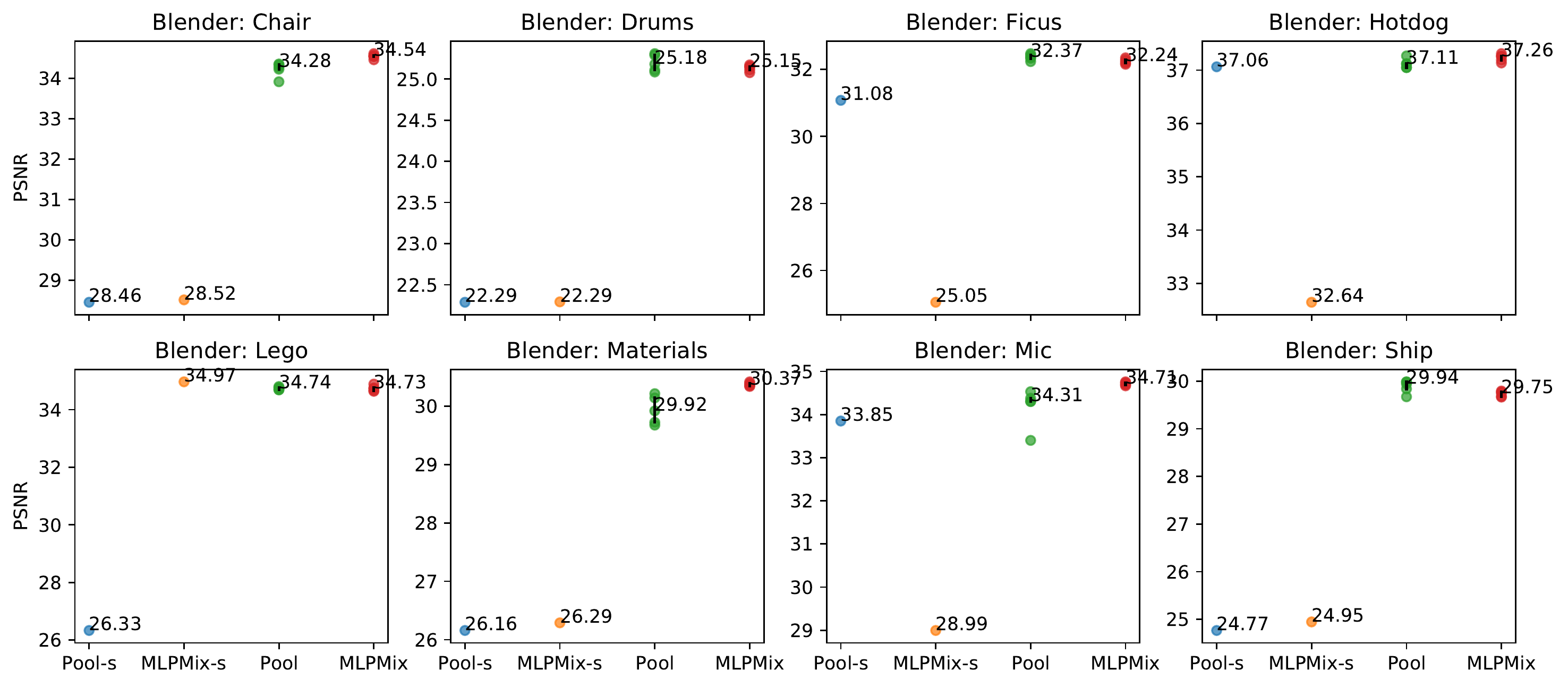} \\
\vspaceScatter
\includegraphics[width=\linewidth]{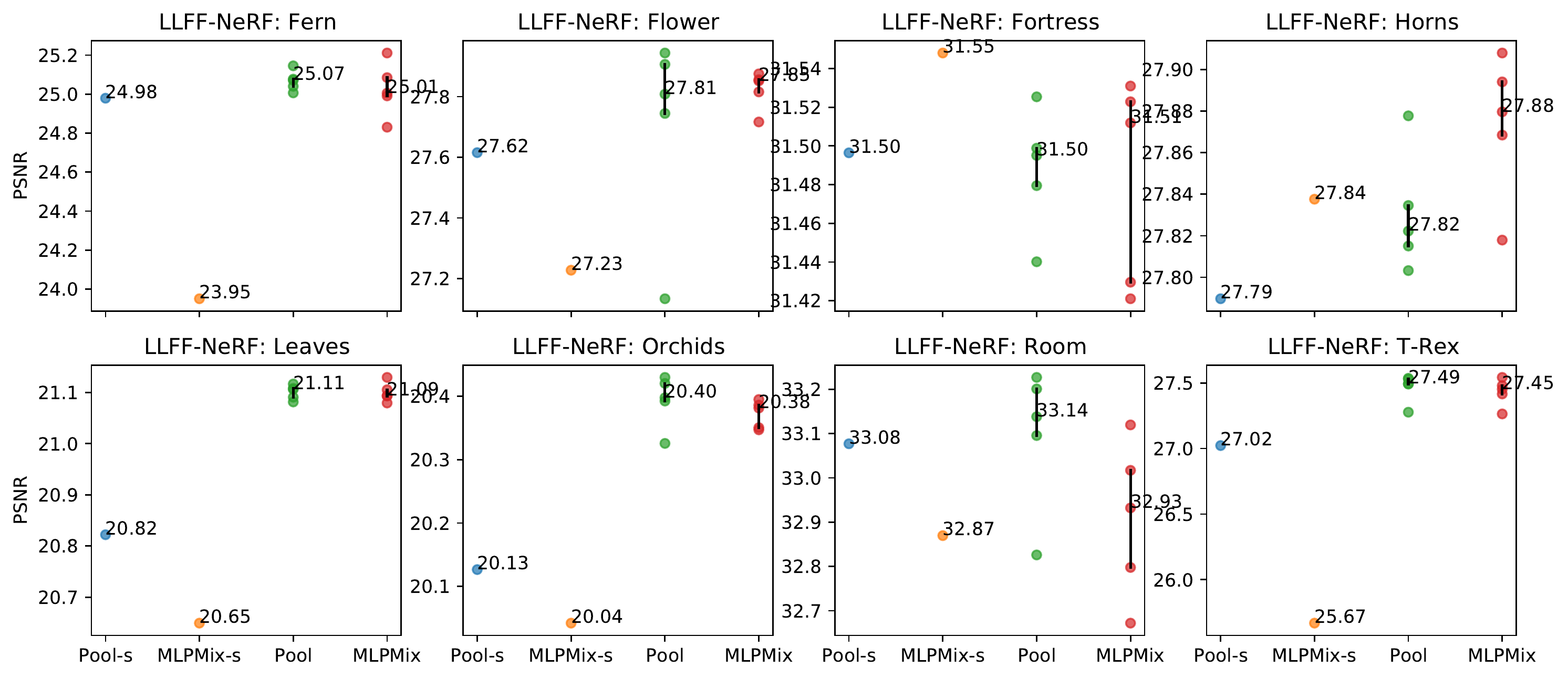}
\caption{{\bf Training from scratch vs.\ two-stage training.}
\capScatter
\emph{Pool} and \emph{MLPMix} are our proposals trained with the
two-stage procedure (\isArXiv{Section~\ref{sec:efftrain}}{Section~2.1}),
while \emph{Pool-s} and \emph{MLPMix-s} are the same proposers but
trained from scratch.
}
\label{fig:res:twostage:psnr}
\end{figure}
}

\newcommand{\capSotaMethods}{
GRF is the method of~\cite{Trevithick20}.
NSVF~\cite{Liu20} only reports results on \emph{Blender}
as the method is not capable of handling unbounded scenes in \emph{LLFF-NeRF}.
IBRNet~\cite{Wang21a} is missing because it does not report
per-scene performances, see \isArXiv{Table~\ref{tab:res:sota}}{Table~2 of the main paper} for
the average results.
}

\newcommand{\figResSotaPsnr}{
\begin{figure}[t]
\centering
\includegraphics[width=\linewidth]{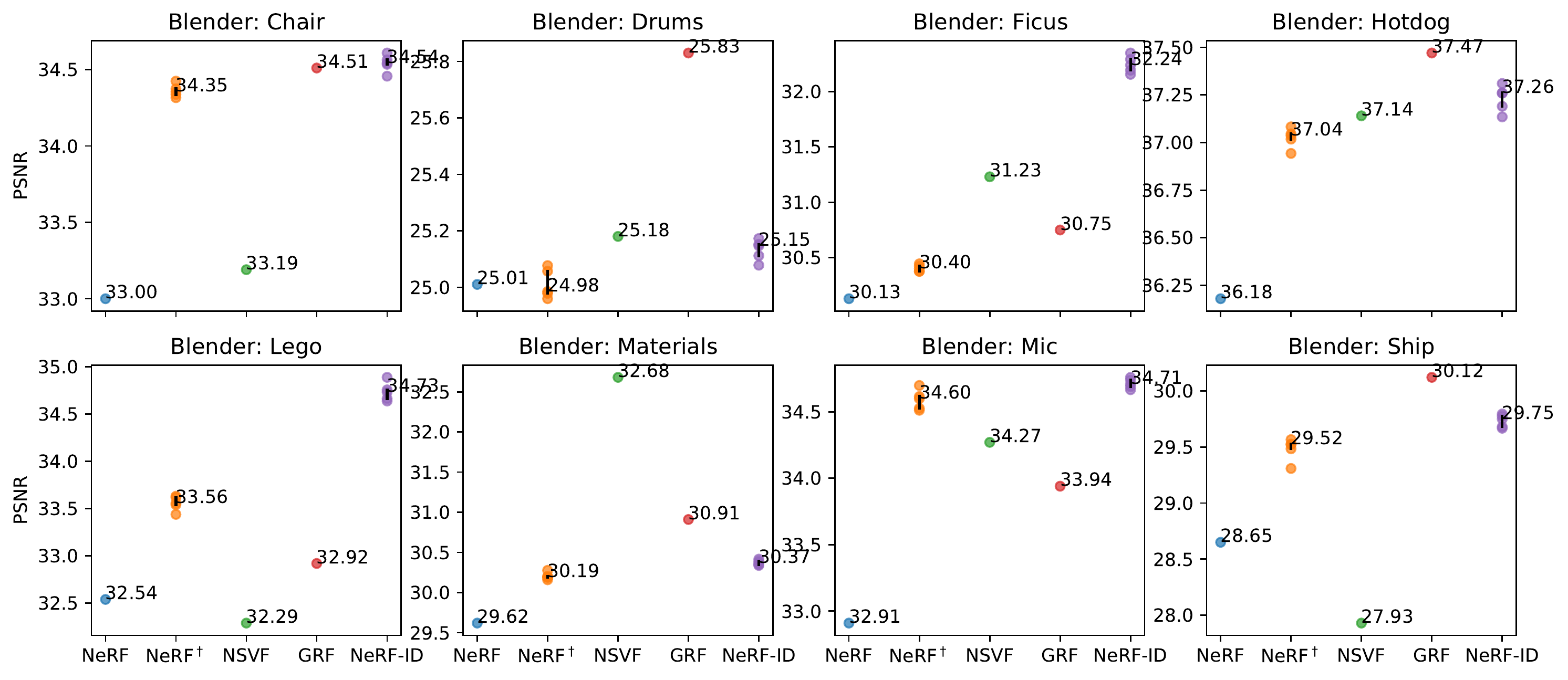} \\
\vspaceScatter
\includegraphics[width=\linewidth]{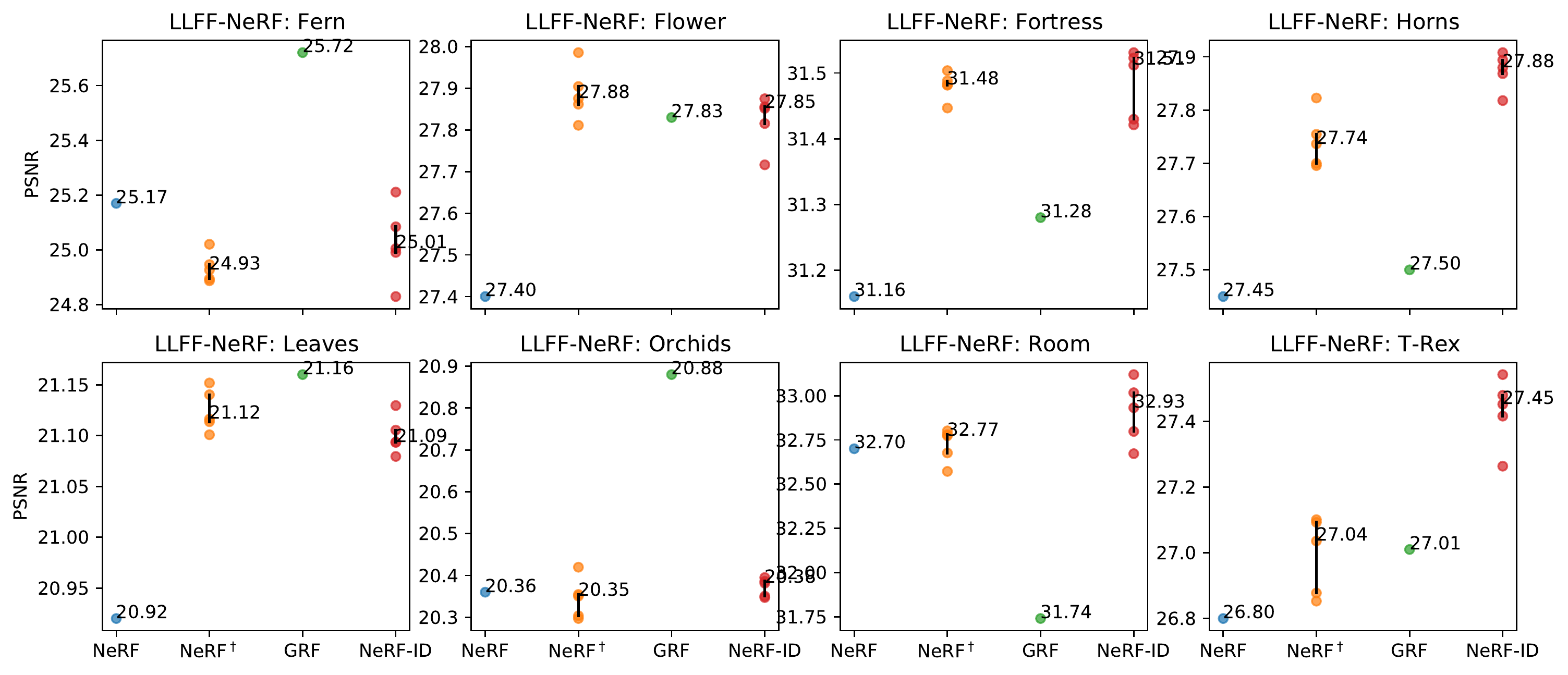}
\caption{{\bf Comparison with the state-of-the-art (PSNR).}
\capScatter
\capNerf
\capSotaMethods
}
\label{fig:res:sota:psnr}
\end{figure}
}

\newcommand{\figResSotaSsim}{
\begin{figure}[t]
\centering
\includegraphics[width=\linewidth]{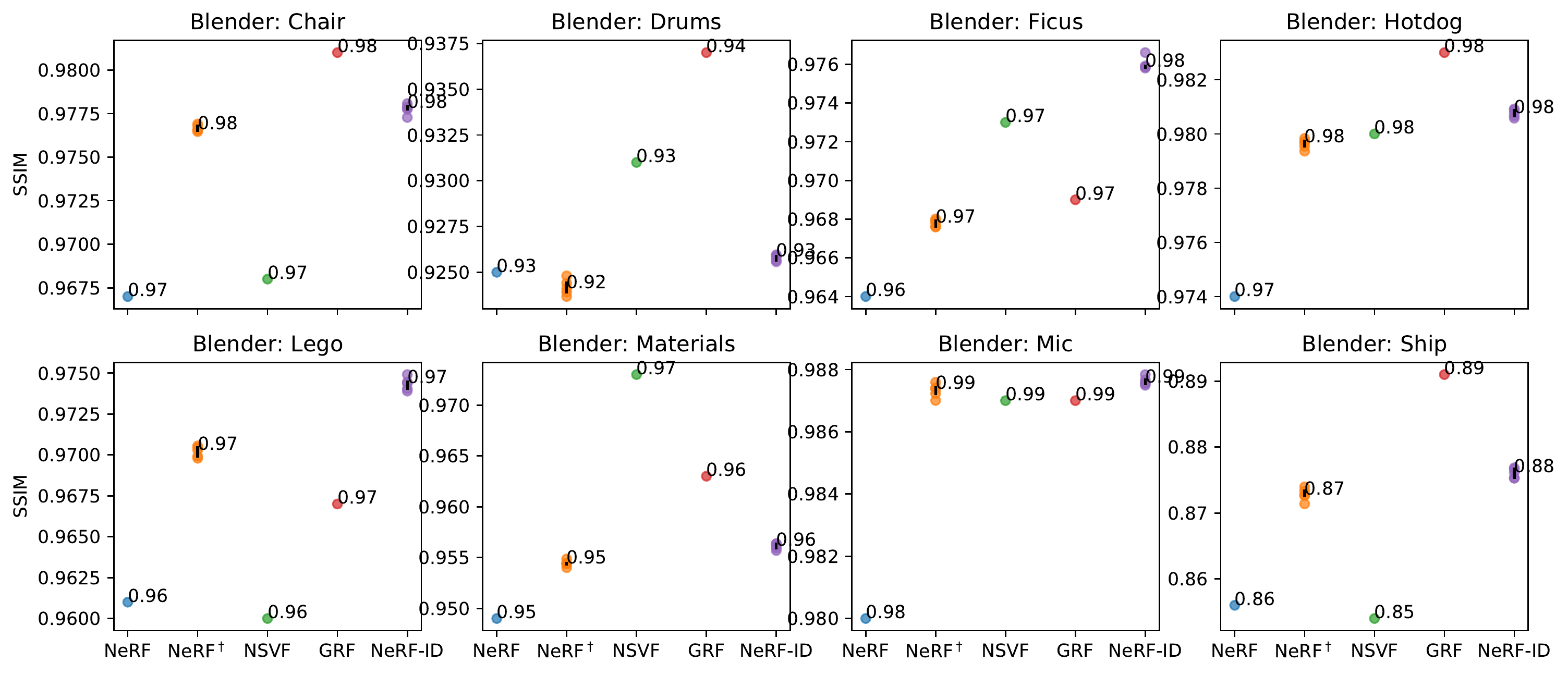} \\
\vspaceScatter
\includegraphics[width=\linewidth]{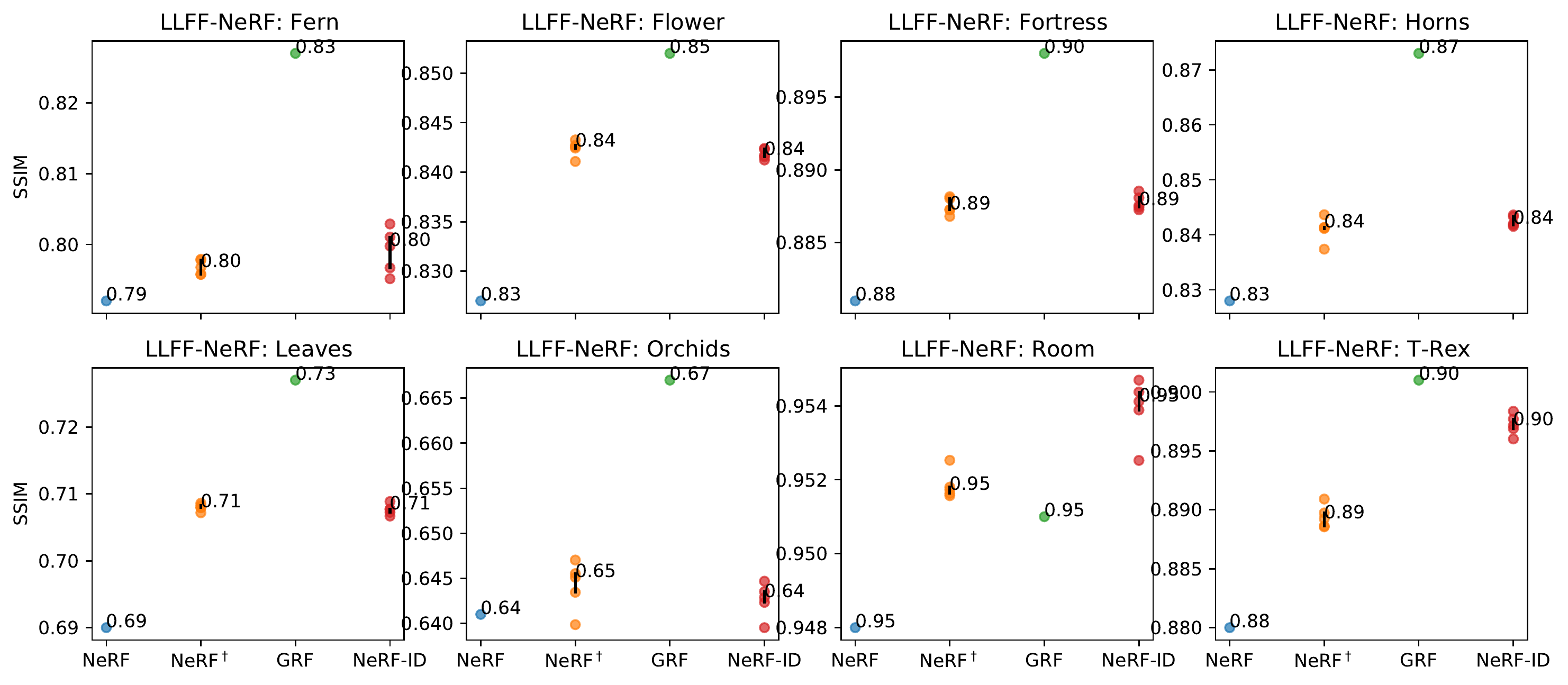}
\caption{{\bf Comparison with the state-of-the-art (SSIM).}
\capScatter
\capNerf
\capSotaMethods
}
\label{fig:res:sota:ssim}
\end{figure}
}

\newcommand{\figSpeedupBlender}{
\begin{figure}[t]
\centering
\includegraphics[width=0.9\linewidth]{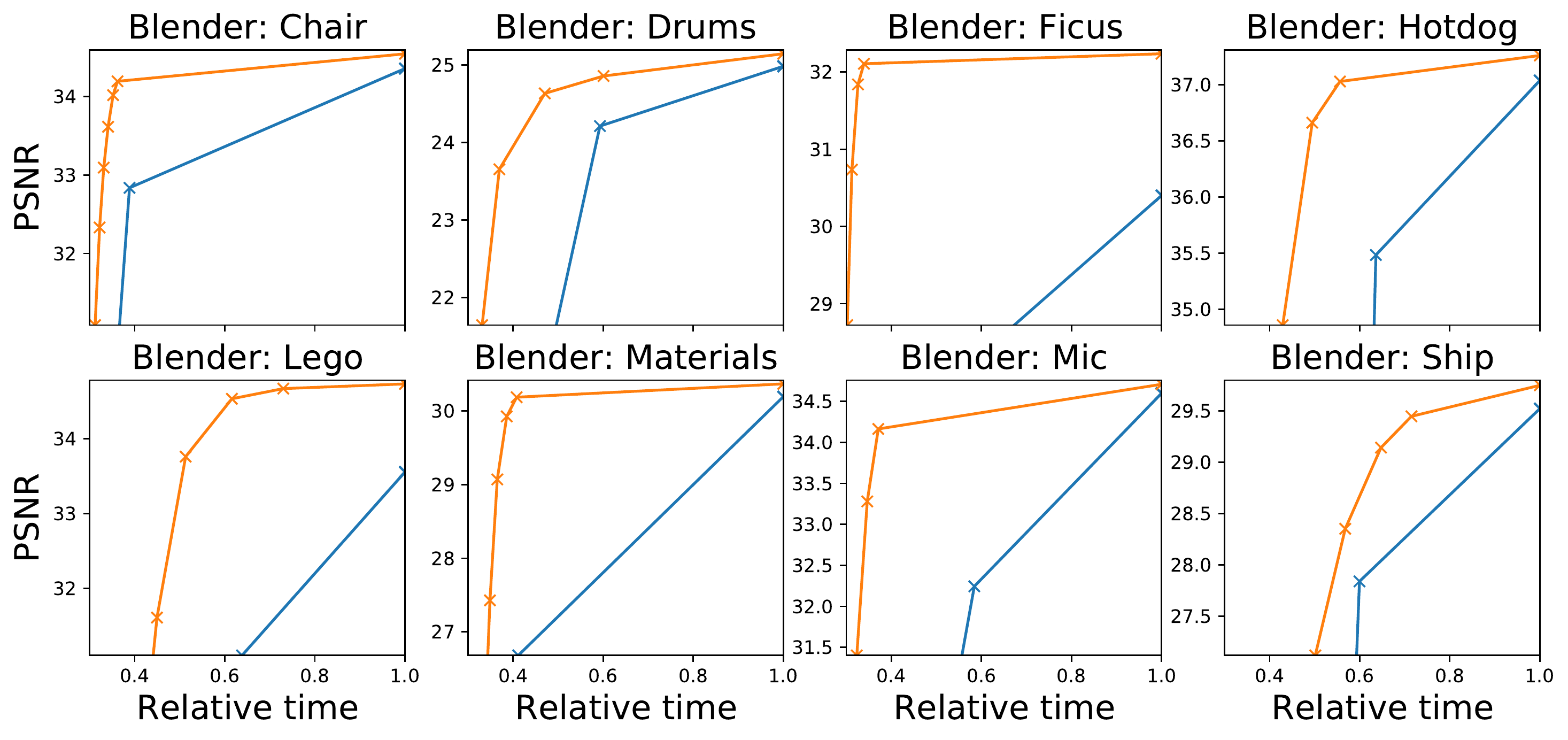}
\caption{{\bf Speedup via importance prediction,
{\color{blue} \NeRFre} vs.\ our {\color{orange} \NeRFour}.}
Samples deemed to be important by the proposer are kept,
different operating points are obtained by varying the importance threshold.
`Relative time' is the time spent on rendering the scene
relative to using all samples.
}
\label{fig:res:speedupblender}
\end{figure}
}

\section*{Appendix overview}

\isArXiv{}{The question ``What is learnt?'' is explored in Appendix~\ref{sec:app:whatlearnt}.}
Appendix~\ref{sec:app:arch} contains full details of the model architectures.
Appendix~\ref{sec:app:training} contains additional details of the training procedure.
Full results of all experiments are provided in Appendix~\ref{sec:app:results}.

\isArXiv{}{
\section{What is learnt?}
\label{sec:app:whatlearnt}

\whatislearnt
}

\section{Architectures}
\label{sec:app:arch}

The architectures of the coarse and fine networks are identical
to the ones in NeRF~\cite{Mildenhall20}.
The proposer architectures are shown in detail in Figure~\ref{fig:archdetail}.

The most expensive operations of
\emph{Pool} and \emph{MLPMix} are the
ones applied on each of the $N_c=64$ coarse features, \ie operations before
the average pooling.
The bottleneck is the single fully connected layer of size $256 \times 256$.
Recall that the NeRF backbone~\cite{Mildenhall20} has 8 of these layers
in each of the coarse and fine networks, and therefore an FC of size
$256 \times 256$ is already being applied on a total of
$8 \times (N_c + N_c + N_f) = 2048$ vectors,
and the additional application on $N_c=64$ vectors by these proposers is negligible.

The \emph{Transformer} proposer is somewhat more expensive due to the lack of
pooling operations, and to make its cost reasonable we constrain it heavily
-- features are projected down to 16-D, single transformer blocks are being used
for the encoder and the decoder, and only a single attention head is being used.

\figArchDetail
\afterpage{\FloatBarrier}

\section{Training details}
\label{sec:app:training}

Here we provide further details to complement
\isArXiv{Section~\ref{sec:res:protocol}}{Section~3.1 of the main paper}.

\paragraph{General training details.}
As with the original NeRF implementation~\cite{Mildenhall20},
there are two slight differences in the setup
for the two datasets -- for \emph{LLFF-NeRF} the color of infinity is taken to be black
and during training random Normal noise is added to the activations
before the ReLU that produces density $\sigma$,
while for \emph{Blender} no noise is added and the color of infinity is white.

\paragraph{Two-stage training.}
Recall from \isArXiv{Section~\ref{sec:efftrain}}{Section~2.1} that the training
is split into two stages of equal duration, where during the first stage
the learnt proposer is trained to mimic the heuristic one,
while in the second stage all components are trained in an end-to-end manner.
When training \NeRFour, for fairness, we use exactly the same settings
including the number of training rays and SGD steps as for \NeRFre.
The only slight difference is that at the start of the second phase
we reset the optimizer state and perform another linear warmup
for 1k steps.
This is to reduce the shock to the system when switching between the
two training regimes, though we still see some undesirable behaviour
(loss rises sharply when switching stages).

As explained in \isArXiv{Section~\ref{sec:efftrain}}{Section~2.1}, during phase 1,
the learnt proposer is trying mimic the heuristic one, which is achieved
by minimizing the L2 matching loss between the two sets of proposals.
Each heuristic sample is matched with
the closest proposal produced by the learnt proposer.
The matching is aggressively greedy for speed reasons as
all other versions (Hungarian algorithm for optimal matching,
greedy 1-1 matching)
significantly slowed down training.

\paragraph{Importance prediction.}
As explained in \isArXiv{Section~\ref{sec:importance}}{Section~2.2},
the importance of each proposal can be predicted.
Note that we place a `stop gradient' operation at the input
to the importance predictor,
\ie the input coarse-network-features are not affected
by training the importance predictor.
We have not ablated this design choice,
but our intuition is that allowing this gradient propagation path
could hurt the overall performance because it could push
the proposer and importance predictor
towards a trivial solution that yields a low importance prediction loss
-- produce bad samples and predict they are unimportant.
Furthermore, using the `stop gradient' also simplifies comparisons
(training the importance predictor does not affect the rest of
the network)
and removes the need for an extra hyper-parameter that
is the weight of the importance predictor loss
(as this loss and the main loss do not interact so Adam~\cite{Kingma15}
scales them independently).

\isArXiv{}{\newpage}
\section{Results}
\label{sec:app:results}

\figResArchPsnr
\figResArchSsim

\paragraph{Proposer architectures.}
Figure~\ref{fig:res:arch:psnr} complements \isArXiv{Table~\ref{tab:res:arch}}{Table~1 of the main paper},
showing PSNRs for all experimental runs.
It confirms that the findings of the main paper are statistically
significant, where \emph{Pool} and \emph{MLPMix} proposers
tend to be the best, but \emph{Pool} has a higher variance
and is significantly worse on two scenes
(\emph{Blender: Materials}, \emph{Blender: Mic}).
Figure~\ref{fig:res:arch:ssim} shows the evaluation in terms
of SSIM, which reveals the same pattern.

\figResMoreArchPsnr

Figure~\ref{fig:res:morearch:psnr} shows additional ablations
on the \emph{Pool} proposer architecture.
The \emph{Blind} proposer, which ignores input coarse-features
and therefore simply always outputs the same proposals,
works badly, showing that it is not sufficient to follow
such a degenerate strategy.
The \emph{No-position} proposer,
which ignores the positions of coarse samples along the ray,
performs badly on \emph{Blender} because it is insensitive
to feature ordering, \eg if produces the same proposals
for two colinear rays of opposite directionality,
but is competitive on \emph{LLFF-NeRF} as all its scenes are
forward-facing.
The exact manner in which positions are incorporated is not so important
--
\emph{Concat} (positional encodings are concatenated with the features)
and
\emph{Learnt-position} (learnt positional encodings are summed with the features)
perform similarly to the \emph{Pool} proposer (sinusoidal positional encodings~\cite{Mildenhall20} are summed with the features).

\figResScratch

\paragraph{Two-stage training.}
Figure~\ref{fig:res:twostage:psnr} shows
that training from scratch sometimes works but often fails,
while two-stage training (\isArXiv{Section~\ref{sec:efftrain}}{Section~2.1}) is effective.

\figResSotaPsnr
\figResSotaSsim

\paragraph{Comparison with the state-of-the-art.}
Figures~\ref{fig:res:sota:psnr} and~\ref{fig:res:sota:ssim}
complement \isArXiv{Table~\ref{tab:res:sota}}{Table~2 of the main paper},
showing PSNRs and SSIMs for all experimental runs and all scenes.
They confirm that the findings of the main paper are statistically
significant.

\figSpeedupBlender

\paragraph{Speedup.}
Figure~\ref{fig:res:speedupblender} shows speedups
on the \emph{Blender} dataset
obtained by
only querying the fine network on samples deemed to be important
by the proposer (\isArXiv{Section~\ref{sec:importance}}{Section~2.2});
\isArXiv{Figure~\ref{fig:res:speedupllff}}{Figure~3 of the main paper} shows the results for \emph{LLFF-NeRF}.
Speedups are especially large for \emph{Blender} because many pixels
correspond to the background, so the corresponding rays can be
heavily pruned.
\NeRFour consistently dominates \NeRFre, producing better images
using fewer computations.
In many cases it is possible to render views 50\% faster while still
achieving better results than \NeRFre.

\isArXiv{}{
{\small
\bibliographystyle{ieee}
\bibliography{bib/shortstrings,bib/more,bib/vgg_local,bib/vgg_other}
}

}{}

\end{document}